\documentclass[letterpaper, 10 pt, conference]{./support/ieeeconf}
\usepackage{times}
\usepackage[pdftex]{graphicx}
\usepackage{subfigure}
\usepackage{amsmath,amssymb,amsopn,amstext,amsfonts}
\usepackage{cancel}
\usepackage[space]{cite}
\usepackage{pdfsync}
\usepackage{balance}
\usepackage{color}
\usepackage{mathtools}
\usepackage{algorithm}
\usepackage{algpseudocode}
\algnewcommand\algorithmicforeach{\textbf{for each}}
\algdef{S}[FOR]{ForEach}[1]{\algorithmicforeach\ #1\ \algorithmicdo}

\usepackage{bm}

\usepackage{diagbox}
\usepackage{float}
\usepackage{epstopdf}
\usepackage{pifont}
\usepackage{fixltx2e}
\usepackage{amsmath}
\usepackage{multirow}
\usepackage{url}
\usepackage[linkcolor=black,citecolor=black,urlcolor=black,colorlinks=true]{hyperref}

\graphicspath{{./figures/}}
\DeclareGraphicsExtensions{.png,.jpg,.eps,.pdf}
\IEEEoverridecommandlockouts
\title{\LARGE \bf FIESTA: Fast Incremental Euclidean Distance Fields for Online Motion Planning of Aerial Robots }
\author{Luxin Han, Fei Gao, Boyu Zhou and Shaojie Shen%
	\thanks{ All authors are with the Department of Electronic and Computer Engineering, Hong Kong University of Science and Technology, Hong Kong, China. {\tt\small $\{$luxin.han, fgaoaa, bzhouai, eeshaojie$\}$@ust.hk}} }

\begin{document}
	
\maketitle
\thispagestyle{empty}
\pagestyle{empty}
\begin{abstract}
Euclidean Signed Distance Field (ESDF) is useful for online motion planning of aerial robots since it can
easily query the distance and gradient information against obstacles. Fast incrementally built ESDF map is the bottleneck for conducting real-time motion planning. In this paper, we investigate this problem and propose a mapping system called \textbf{FIESTA} to build global ESDF map incrementally. By introducing two independent updating queues for inserting and deleting obstacles separately, and using Indexing Data Structures and Doubly Linked Lists for map maintenance, our algorithm updates as few as possible nodes using a BFS framework. Our ESDF map has high computational performance and produces near-optimal results.  We show our method outperforms other up-to-date methods in term of performance and accuracy by both theory and experiments. We integrate FIESTA into a completed quadrotor system and validate it by both simulation and onboard experiments. We release our method as open-source software for the community\footnote[1]{\url{https://github.com/hlx1996/FIESTA}}. 
\end{abstract}

\section{Introduction}
\label{sec:introduction}

For a fully autonomous Micro Aerial Vehicle (MAV), the perception-planning-control pipeline takes environmental measurements as input and generates control commands. The mapping module, which provides a foundation for onboard motion planning, is one of the most essential components in this system.

\begin{figure}[t]
	\centering
	\subfigure[  ]
	{\includegraphics[width=0.9\columnwidth]{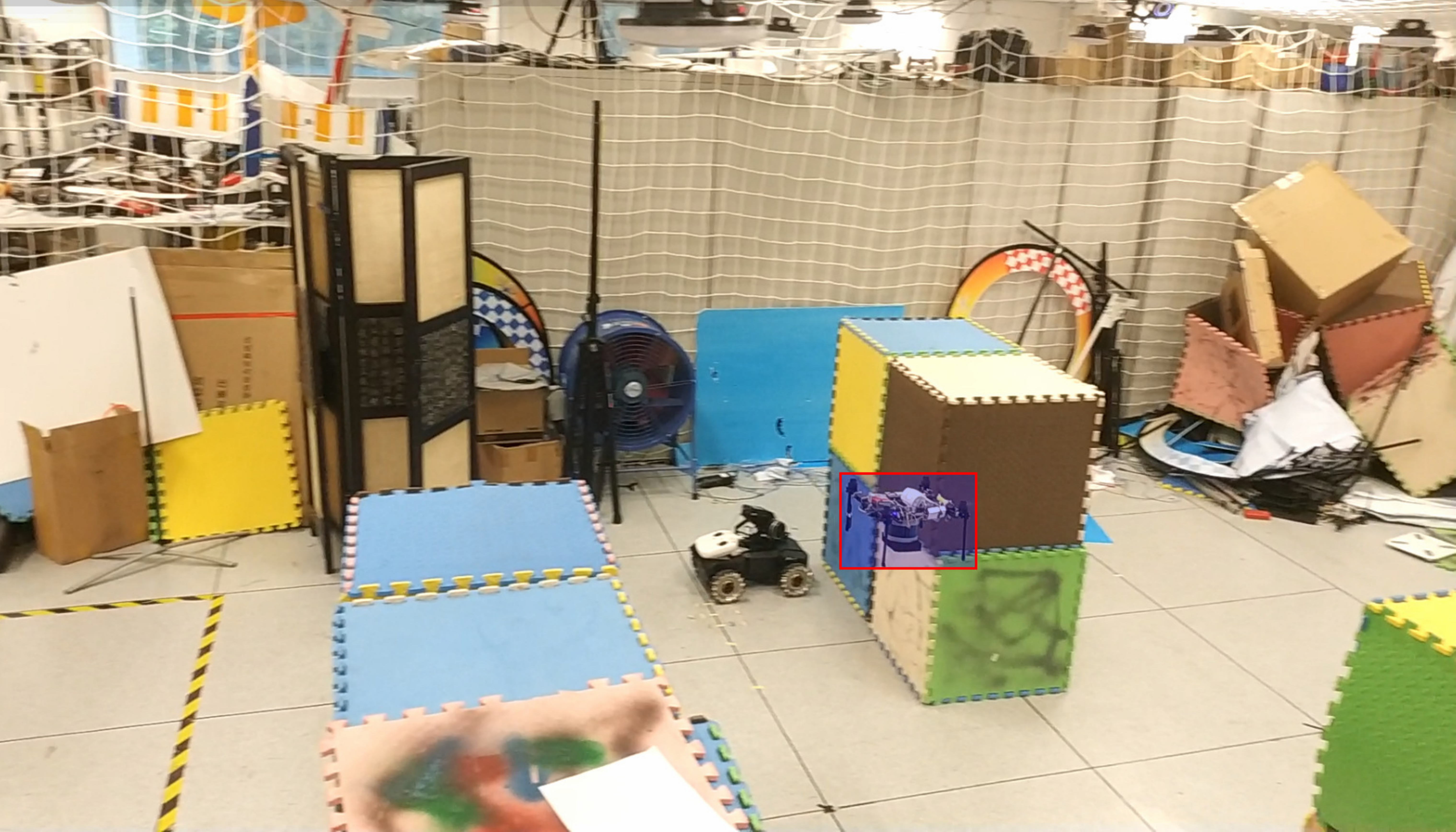}}
	\subfigure[]
	{\includegraphics[width=0.9\columnwidth]{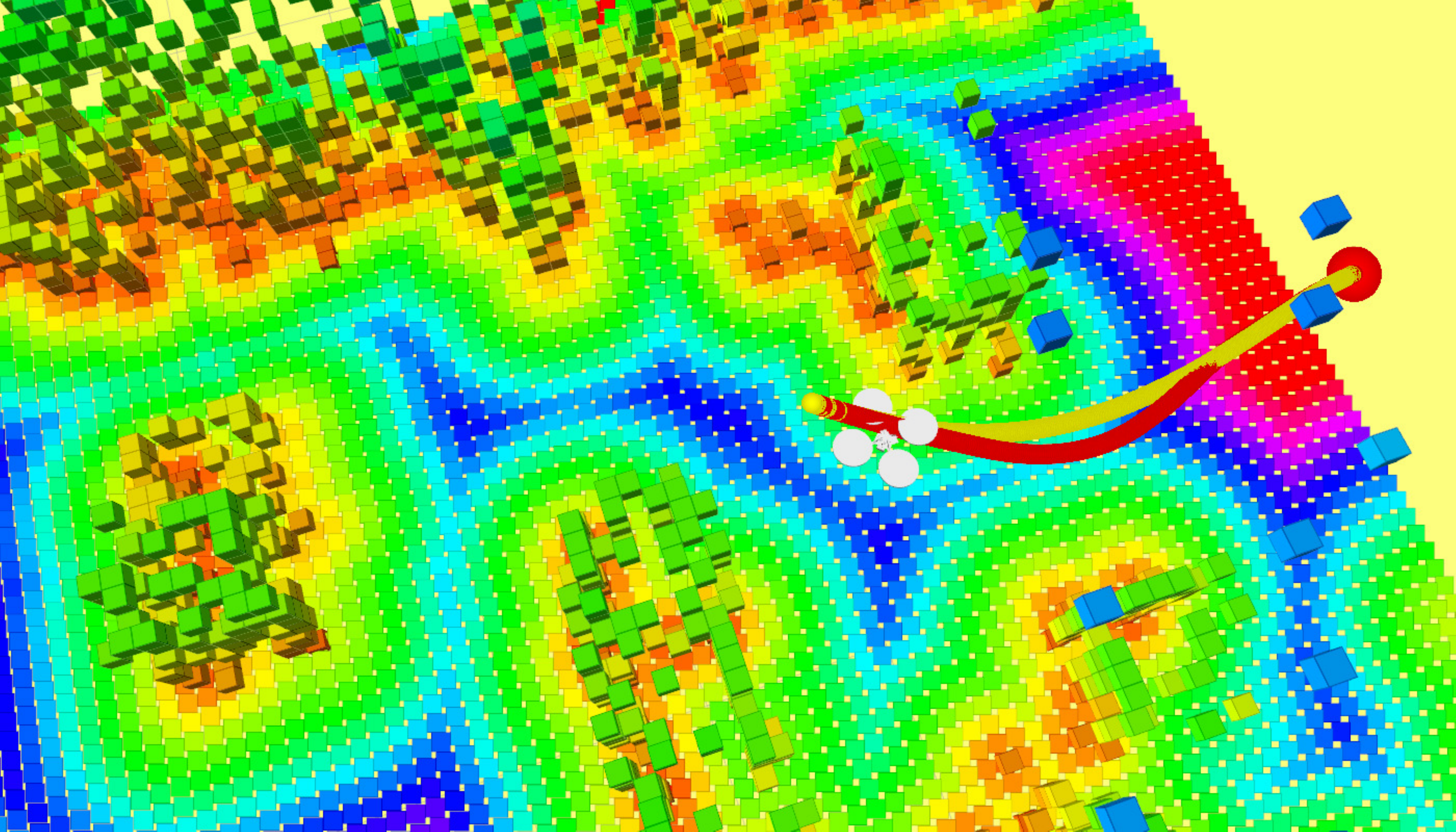}}
	\caption{\label{fig:realworld} Motion planning experiment using FIESTA for calculate ESDF Map. The quadrotor plans to fly to a designated destination through the maze. \cite{boyu2019ral} is used for motion planning. (a) shows the quadrotor in the experiment, (b) is the corresponding visulization of FIESTA, where cubes are obstacles, plane is  one slice of ESDF map, yellow and red lines are the planned trajectory. Details about the experiments are shown in this video\protect\footnotemark[2].}
\vspace{-0.5cm}
\end{figure} 
\footnotetext[2]{\url{https://www.youtube.com/watch?v=pgRi8LOnT6Y}}
There are many well-developed data structures and algorithms for mapping. The basic requirement of a mapping system is to balance the accuracy of fusing depth measurements and the overhead of storing a fine representation of the environment. Representative mapping frameworks include Octomap~\cite{wurm2010octomap} and TSDF(Truncated Signed Distance Field)~\cite{newcombe2011kinectfusion}, etc. However, for the purpose of autonomous quadrotor navigation, what is truly useful is the information of free space, instead of obstacles. A desirable map for planning must have the capability of fast querying free/occupied status, or getting the distance information to obstacles, such as the Euclidean Signed Distance Field (ESDF) map. 

ESDF map has the advantage to evaluate the distance and gradient information against obstacles and is, therefore, necessary for gradient-based planning method, such as CHOMP~\cite{ratliff2009chomp}. The gradient-based method tends to push the generated trajectories away from obstacles to improve path clearance. Therefore it is particularly useful in quadrotor planning as presented in~\cite{oleynikova2016continuous},~\cite{fei2017iros} and~\cite{usenko2017real}. For real-time onboard motion planning, the efficiency and accuracy in maintaining and updating the ESDF map is always the bottleneck. In~\cite{felzenszwalb2012distance}, an efficient method using distance parabolic curve is proposed to compute the ESDF globally. However, for quadrotor platform where the sensing range and onboard computing resource are both very limited, incrementally updating the ESDF map is demanding for real-time onboard planning. Another way to update ESDF map in real-time is to keep and maintain only a small local map sliding with the quadrotor, such as presented in~\cite{usenko2017real}. Although this compromise works in quadrotor local (re-)planning, it discards all past map information and can not be used for applications where global or repeatable planning is needed. 

Voxblox~\cite{oleynikova2017voxblox} are also proposed to incrementally build the ESDF map. It computes the ESDF map directly from a Truncated Signed Distance Field (TSDF) map. It leverages the distance information already contained within the truncated radius in the TSDF map and then expands to all the voxels in the map using BFS (Breadth-First Search). It can work in real-time on a dynamically growing map, however, there are actually two categories of errors existing in the final ESDF map. Firstly, it conducts the BFS and updates the ESDF according to the quasi-Euclidean distance. Therefore the computed distance may have larger error compared to the actual Euclidean distance. Secondly, it relies on TSDF-based mapping, but the TSDF projective distance may overestimate the actual Euclidean distance to the nearest surface.

In this paper, we propose \textbf{FIESTA} (\textbf{F}ast \textbf{I}ncremental \textbf{E}uclidean Di\textbf{STA}nce Fields), which is a lightweight and flexible mapping framework for building ESDF map incrementally. Our method takes the pose estimations and depth measurements as input, and updates the ESDF map globally with the minimal computational overhead. This is done by our elaborately designed data structures and ESDF updating algorithm. Our data structures build the foundation for our high-efficiency ESDF updating algorithm, and also provides an option to trade-off the time complexity and space complexity. Our proposed algorithm expands as few as possible nodes using a BFS framework and results in much more accurate ESDF values. What's more, unlike Voxblox~\cite{oleynikova2017voxblox} depending on TSDF, our algorithm doesn't rely on any specific types of mapping framework. It means that our method can be applied to any general mapping framework, including occupancy grid map and TSDF-based map. The contributions of this work are:
\begin{itemize}
	\item Elaborately designed data structures for fast updating the ESDF map incrementally.
	\item A novel ESDF updating algorithm which expands as few as possible nodes and obtains near-optimal results. 
	\item Theoretical and practical analysis of time and space complexity, accuracy, and optimality.
	\item Integration of the proposed ESDF map into a completed quadrotor system and demonstrations of onboard MAV motion planning.
	\item Release the proposed mapping framework as open-source software.
\end{itemize}

\section{Related Work}
\label{sec:RelatedWork}
There are a lot of different mapping frameworks used in quadrotor planning. Some of them are derived from a general map used in robotic perception. In~\cite{fei2018icra}, a grid map is used for fast collision checking and occupancy evaluation. Using the occupancy map, a field-based path searching is conducted, followed by a hard-constrained piecewise trajectory generation. In ~\cite{CheLiuShe2016}, the authors utilize the Octomap to group large free space for searching a collision-free path and generating safe trajectories. Moreover, point clouds are directly used for planning in~\cite{fei2018jfr}, where the safe space of the environment is extracted by randomly querying the nearest neighbor using a Kd-tree of the point cloud. Some other mapping frameworks tailored for robotic navigation are also proposed recently. Topomap~\cite{blochliger2018topomap} builds convex clusters based on sparse features points from a visual SLAM system. The authors build a sparse topological graph using these convex clusters, where a path can be found very efficiently. Sparsemap~\cite{oleynikova2018sparse} shares a similar idea with Topomap. The authors propose a complete pipeline to extract a 3D Generalized Voronoi Diagram (GVD) based on an ESDF and obtain a thin skeleton diagram representing the topological structure of the environment. MAV planning is conducted in this topological map. Nanomap~\cite{florence2018nanomap} propose a novel map structure, which can be built and queried range search efficiently. In Nanomap~\cite{florence2018nanomap}, the authors propose to keep only recently depth raw measurements to do collision checking, instead of explicitly maintaining a fused map. Fast quadrotor flight is supported by this map structure. 

ESDF map is widely used in gradient-based robotics motion planning, where the distance and gradient information to obstacles are necessary. CHOMP~\cite{ratliff2009chomp} uses the method~\cite{felzenszwalb2012distance} to compute the ESDF map. It starts with a deterministic occupancy grid map, which means the value is either 0 or 1, and compute the EDT (Euclidean Distance Transform) for the map. Computing the EDT using~\cite{felzenszwalb2012distance} is efficient. However, motion planning suffers from repeating calculating ESDF value again and again. It can't satisfy both the map accuracy and updating frequency, which in return, limits the performance of the planning module running onboard.

The most relevant work to ours is Voxblox~\cite{oleynikova2017voxblox}, which also focus on building the ESDF map incrementally. Voxblox firstly integrates sensor data into a TSDF Map and then calculates ESDF from TSDF using a BFS(Breadth-First-Search) framework. Voxblox fully utilizes the distance information in the TSDF map, has real-time performance on CPU and allows dynamically-growing map size. However, this approach has some drawbacks. Firstly, the ESDF values are far away from accurate. There are actually two categories of errors existing in the final ESDF map. (i) It uses BFS to update the ESDF information of neighbor points and updates them with the length of a broken line, which is also called quasi-Euclidean expanding. In this way, the final ESDF is not the real Euclidean distance, which may have larger error compared to the actual Euclidean distance. (ii) It relies on TSDF-based mapping, where the TSDF projective distance may overestimate the actual Euclidean distance to the nearest surface. Instead, our work incrementally calculates ESDF directly from an occupancy grid map, one of the simplest mapping data structures, which eliminates error from (ii) fundamentally. As for (i), we also use BFS but updates its neighbors in another way. It is not accurate, either, as we prove that in Section~\ref{sec:Optimality} that all ESDF updating algorithms based on BFS cannot be accurate. However it is much better than a quasi-Euclidean way in an order of magnitude in both theory and practice, which will be shown in Section~\ref{sec:Optimality} and Section~\ref{sec:DatasetResults}, respectively. Secondly, Voxblox highly depends on TSDF mapping, which results in its high complexity and low extensibility. Its complexity also makes it hard to be transplanted into other applications. And its updating starts from changes of TSDF voxels, makes the whole algorithm hard to analyze and optimize. On the contrary, our proposed system is a lightweight solution which introduces two independent updating queues for inserting and deleting obstacles separately, gives a natural explanation of the starting points of the BFS, makes it easier to analyze and optimize. 

\section{System Framework}
\label{sec:SystemFramework}
\begin{figure}[t]
	\centering
	\includegraphics[width=1\columnwidth]{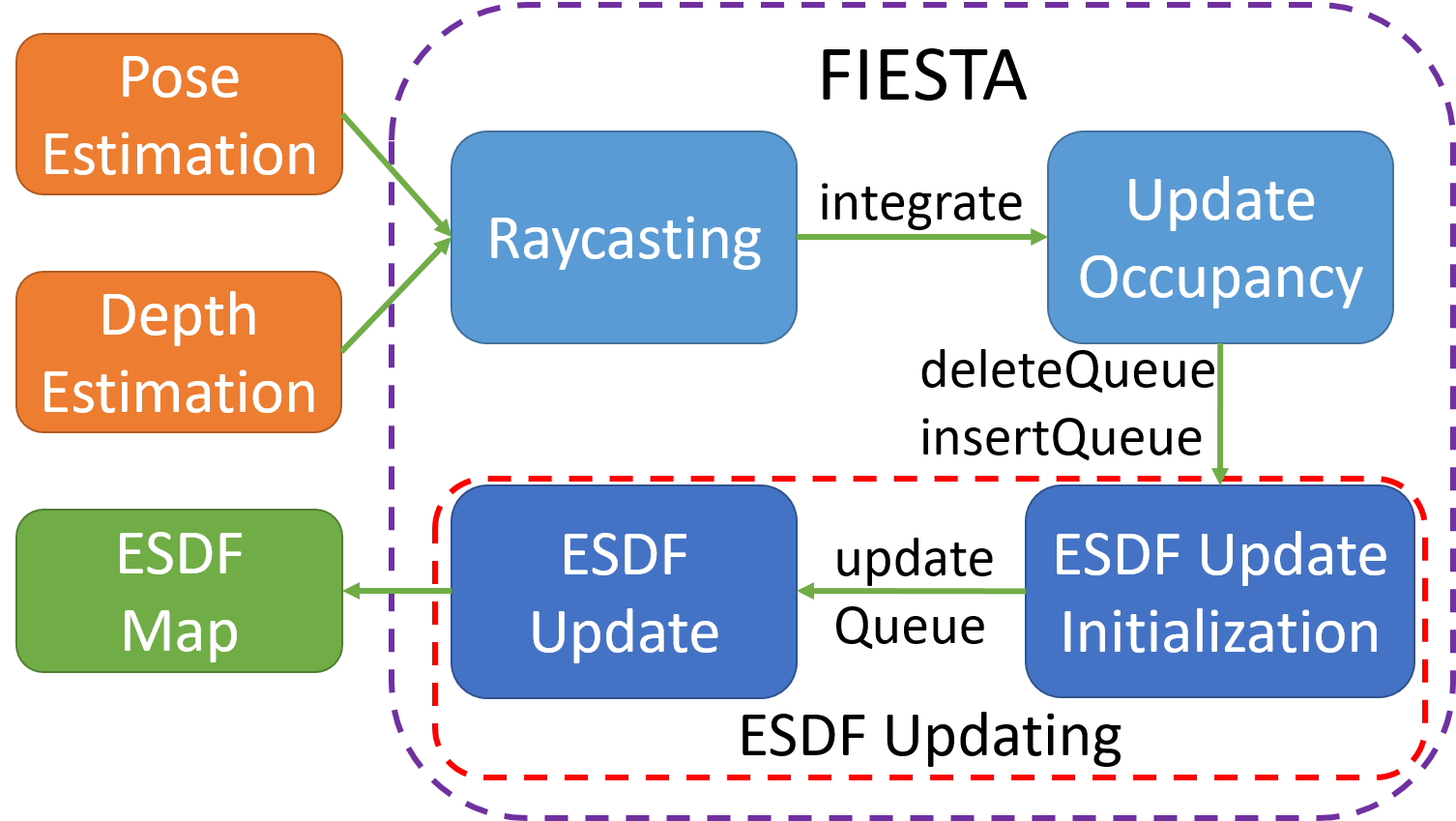}
	\caption{ System overview of our proposed ESDF mapping framework.
		\label{fig:SystemFramework}}
	\vspace{-0.7cm}
\end{figure}

The overview of our system FIESTA is shown in Fig.~\ref{fig:SystemFramework}. Firstly, we get depth measurements from stereo, RGB-D sensors or monocular depth estimation, and pose measurements from external devices such as GPS and Vicon, or internal pose estimation such as VIO (Visual-Inertial Odometry). Then we use \textbf{raycasting} to integrate them into the occupancy grid map, which is the data structure we choose to store the occupancy information in \textbf{FIESTA}. In this process, all voxels that change their occupancy status are added into two queues named $insertQueue$ and $deleteQueue$ respectively. After that we merge these two queues into one queue named $updateQueue$ by a procedure called \textbf{ESDF Update Initialization} and update all the voxels that ESDF may change using ESDF Update Algorithm based on BFS.

\section{Data Structures}
\label{sec:DataStructure}
This section describes all data structures used in FIESTA. Occupancy Grip Map is used to integrate occupancy information. Voxel Information Structures store all information with respect to voxels. Indexing Data Structure is the mapping from a voxel coordinate towards corresponding Voxel Information Structure. And Doubly Linked Lists are used especially for updating ESDF efficiently when an occupied voxel becomes free.

\subsection{Occupancy Grid Map \& Voxel Information Structure}
\begin{table}[t]
	\centering
\begin{tabular}{|l|l|l|}
	\hline
	Name & Meaning & Abbreviation \\ \hline
	position & voxel coordinate & $pos$ \\ \hline
	occupancy & probability of occupancy & $occ$ \\ \hline
	ESDF & \begin{tabular}[c]{@{}l@{}}Euclidean distance\\ to the closest obstacle\end{tabular} & $dis$ \\ \hline
	\begin{tabular}[c]{@{}l@{}}Closest Obstacle\\ Voxel Coordinate\end{tabular} & \begin{tabular}[c]{@{}l@{}}the voxel coordinate\\ of the closest obstacle\end{tabular} & $coc$ \\ \hline
	observed & \begin{tabular}[c]{@{}l@{}}whether this voxel\\ is ever observed\end{tabular} & $obs$ \\ \hline
	prev, next, head & used in DLLs & $prev$, $next$, $head$ \\ \hline
	\begin{tabular}[c]{@{}l@{}} Doubly Linked \\List (method)\end{tabular} & \begin{tabular}[c]{@{}l@{}}all voxels that the  closest\\ obstacle is this voxel\end{tabular} & $dll$ \\ \hline
	\begin{tabular}[c]{@{}l@{}}Neighhoods\\ (method)\end{tabular} & \begin{tabular}[c]{@{}l@{}}all observed neighbors\\ of this voxel\end{tabular} & $nbrs$ \\ \hline
\end{tabular}
	\caption{Members and Methods of a VIS \newline and abbreviation in pseudocode}
	\label{tab:Structure}
	\vspace{-1.0cm}
\end{table}

Occupancy Grid Map is used in our method to store the probability of occupancy of voxels. It keeps integrating new occupancy information data obtained from raycasting when a new depth measurement is given. The probabilistic occupancy information is stored in a Voxel Information Structures (VIS) of voxels. Besides that, there are some other important members in VIS. Table~\ref{tab:Structure} lists all useful members with their meanings and abbreviations in the pseudocode of this paper. The last two rows in it are two methods of a voxel. $dll$ will be described in Section~\ref{sec:DLL}, and $nbrs$ means all observed neighbors, with the designated connectivity, such as 6-connectivity or 26-connectivity.

\subsection{Indexing Data Structure}
\label{sec:indexing}
Depends on whether the bounding box of the planning area is pre-known or not and the memory reserved for this algorithm, an array or a hash table is used for indexing work, i.e. mapping a voxel coordinate towards the corresponding pointer of VIS.

In detail, if the bounding box for the planning area is pre-known and the memory is large enough, pointers of VIS of all voxels can be simply put into an array. In this way, when a voxel is queried by its coordinate, we just need to compute the index of it inside that array and then return the corresponding pointer of VIS. On the contrary, if the memory is inadequate or the bounding box is unknown, a hash table is required to convert a voxel coordinate to its corresponding pointer of VIS. With this method, memory consumption is minimized, because all voxels in this data structure have been observed. However, since the number of look-up operations in a hash table is enormous, this will make the performance much worse than the array implementation.

Thus we give an operatable trade-off of these two by making $(block\_size)^3$ voxels to a block, just like what Voxel Hashing~\cite{niessner2013real} did. Hash table here is only used to manage blocks. After we calculate block coordinate from voxel coordinate, that hash table is used to find the corresponding block. All pointers of VIS of all voxels in that block are stored in an array with respect to that block. Because the size of the hash table for blocks is much smaller than that for voxels, the performance can be improved by more usage of memory. We can even use bitwise operation to speed up the converting work, by using $block\_size$ as an integer power of 2, such as 4, 8, 16 etc.

No matter which data structure is used, the time complexity for look-up operation is always $\Theta(1)$ on average. Though the speed-up factor of different data structures is only constant-level, it is crucial for systems requiring real-time performance on limited resources.

\subsection{Doubly Linked Lists}
\label{sec:DLL}
\begin{figure}[t]
	\centering
	\includegraphics[width=1\columnwidth]{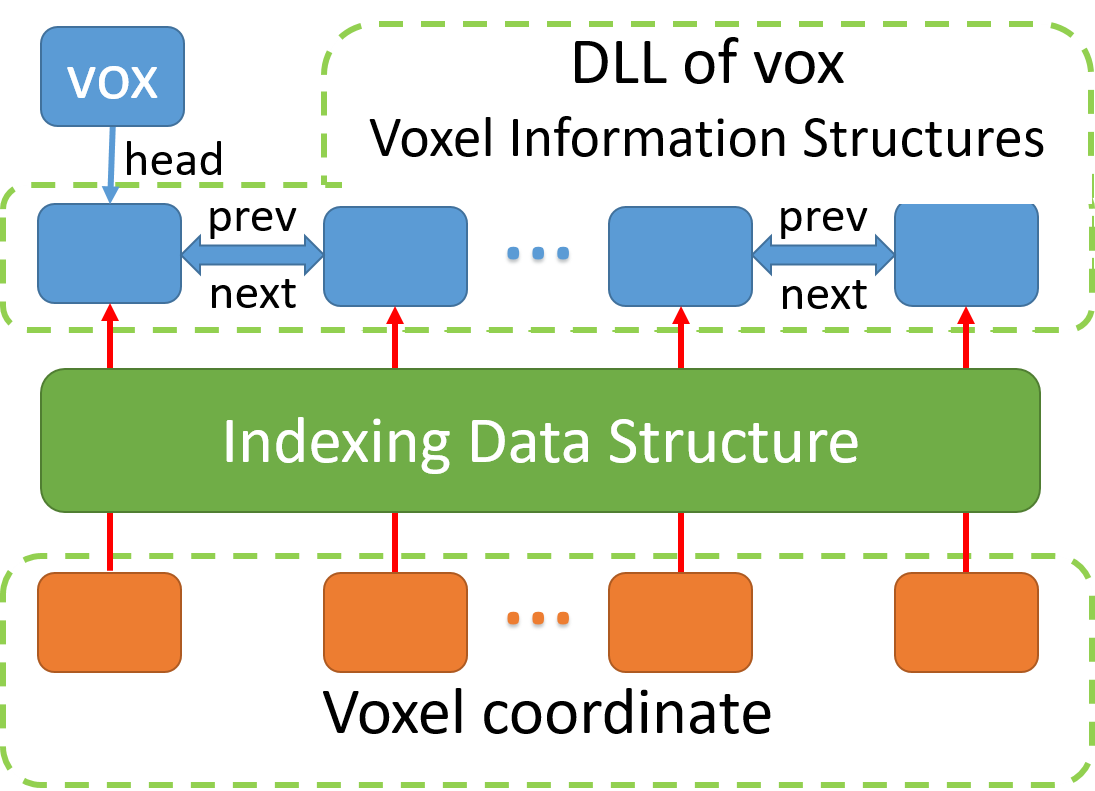}
	\caption{ Diagram among differnt data structures: Indexing Data Structure, Voxel Information Structures and Doubly Linked Lists.
		\label{fig:DLL}}
	\vspace{-0.5cm}
\end{figure}

\textbf{Doubly Linked Lists(DLLs)} are used especially for updating ESDF efficiently when an occupied voxel becomes free (more details in Section~\ref{sec:deleteObstacles}). A DLL is a linked data structure that consists of a set of sequentially linked nodes, with two link fields, reference to the previous and the next node. We use $dll$ of a voxel $vox$, i.e. $dll$ method in Table~\ref{tab:Structure}, to represent all voxels whose closest obstacle is $vox$. They are linked into a DLL, and the head pointer of this DLL can be obtained by the $head$ member of VIS of $vox$. The $dll$ method is actually iterating the whole DLL from the head pointer. Fig.~\ref{fig:DLL} shows the relationship among Indexing Data Structure, Voxel Information
Structures and Doubly Linked Lists.

To the beginning, the closest obstacles of all voxels are initialized to \textbf{Ideal Point}, namely Point at Infinity. We use $\mathcal{IP}$ to represent it. When a voxel is newly observed (more details in Section~\ref{sec:NewlyObserved}), it will be added to a $dll$ of $\mathcal{IP}$. As a consequence, all voxels that are newly observed but not yet updated are linked in the $dll$ of $\mathcal{IP}$.

With the help of the Indexing Data Structure and the DLLs, insert into or delete from a DLL is easily implemented in $\Theta(1)$, because (1) find corresponding VIS using Index Data Structure is $\Theta(1)$, and (2) insert or delete a given node in DLL is $\Theta(1)$. We use \textsc{insertIntoDLL} and \textsc{deleteFromDLL} to denotes these two procedures. They have two parameters, which are the header of DLL and the voxel need to be inserted or deleted.

\section{Algorithms}
\label{sec:Algorithm}

Since ESDF value can be easily calculated by the difference of EDF (Euclidean Distance Field) values for both occupancy grid map and its logical complement\cite{ratliff2009chomp} within the double time of the EDF calculation. In this section, we only explain how to calculate an EDF map. To avoid confusion of notations, we still use ESDF to represent the map we build. 

\subsection{Occupancy Integration}
\label{sec:OccupancyIntegration}
Every time we get a pose estimation and a depth estimation with aligned timestamp, a point cloud of obstacles can be obtained. We use raycasting method to integrate the new occupancy information into the Occupancy Grid Map. After that, all voxels that becomes newly observed are inserted into $dll$ of $\mathcal{IP}$ and marked observed. All voxels that change their occupancy state to occupied or unoccupied are added into two queues named $insertQueue$ and $deleteQueue$ respectively.

\subsection{ESDF Updating Algorithm for Insert-only Case}
\label{sec:InsertOnly}

The ESDF Updating Algorithm is based on a BFS algorithm. To explain it easier, a simplified case where obstacles can only be inserted only is explained firstly. It is the case using a deterministic occupancy grid map where all occupancy integrations only make some voxels become occupied by obstacles. In this case, $deleteQueue$ is empty, we simply add everything in $insertQueue$ into $updateQueue$ after simple initialization (see Alg.~\ref{alg:covert} Line~1~-~9). We give a not accurate but useful assumption (see more in Section~\ref{sec:Optimality}) that a new obstacle will only influence ESDF and the closest obstacle of voxels in a continuous (with respect to the designated connectivity) bounded area. Under this assumption, BFS is used to update neighbors. The complete ESDF Updating Algorithm is described in Alg.~\ref{alg:Framework}, where line~5~-~11 is the Euclidean updating way we proposed, like what SDF\_Tools~\cite{UMARMLab2014} did.

\begin{algorithm}[t]
	\caption{ ESDF Updating Algorithm }
	\label{alg:Framework}
	\begin{algorithmic}[1]
		\Require
		$updateQueue$ is the queue from ESDF Updating Initialization;
		\Ensure All voxels need to be updated should be updated with the minimal computational overhead.
		\While {\textbf{not} $updateQueue$.\Call{empty}{\null}}
			\State {$cur$ $\gets$ $updateQueue$.\Call{front}{\null}}
			\State {$updateQueue$.\Call{pop}{\null}}

			\ForEach{$nbr \in cur.nbrs$}
				\If {\Call{dist}{$cur.coc, nbr.pos$} $< nbr.dis$}
					\State {$nbr$.$dis$ $\gets$ \Call{dist}{$cur.coc$, $nbr.pos$}}
					\State {\Call{deleteFromDLL}{$nbr.coc$, $nbr.pos$}}
					\State {$nbr.coc$ $\gets$ $cur.coc$}
					\State {\Call{insertIntoDLL}{$nbr.coc$, $nbr.pos$}}
					\State {$updateQueue$.\Call{push}{$nbr$}}
				\EndIf
			\EndFor
		\EndWhile
	\end{algorithmic}
\end{algorithm}

\subsection{ESDF Updating Algorithm for Fully Dynamic Case}
\label{sec:deleteObstacles}
In the fully dynamic case, obstacles can be either inserted or deleted. They are in the two queues named $insertQueue$ and $deleteQueue$ provided by Section~\ref{sec:OccupancyIntegration}. The measure to deal with $insertQueue$ remains the same. For each voxel in $deleteQueue$, all voxels from the $dll$ of it will be iterated. Status of each term will be cleared back to the observed but not yet updated status, i.e. set the closest obstacle to $\mathcal{IP}$ and set ESDF to $\infty$. After that, try to update their ESDF by existing closest obstacles of their neighbors. If ever updated, insert this voxel to the DLL of its closest obstacle and add it into $updateQueue$. The complete algorithm that merges $insertQueue$ and $deleteQueue$ to $updateQueue$ is shown in Alg.~\ref{alg:covert}. After that, the ESDF Updating Algorithm in Alg.~\ref{alg:Framework} is executed. 

\begin{algorithm}[t]
	\caption{ ESDF Updating Initialization }
	\label{alg:covert}
	\begin{algorithmic}[1]
		\Require
		$insertQueue$ is all voxels changed to occupied.
		
		$deleteQueue$ is all voxels changed to unoccupied.
		\Ensure
		After Alg.~\ref{alg:Framework} using given $updateQueue$, all voxels need to be updated should be updated
		\While {\textbf{not} $insertQueue$.\Call{empty}{\null}}
			\State {$cur$ $\gets$ $insertQueue$.\Call{front}{\null}}
			\State {$insertQueue$.\Call{pop}{\null}}
			\State {\Call{deleteFromDLL}{$cur.coc$, $cur.pos$}}
			\State {$cur.coc$ $\gets$ $cur.pos$}
			\State {$cur.dis$ $\gets$ $0$}
			\State {\Call{insertIntoDLL}{$cur.coc$, $cur.pos$}}
			\State {$updateQueue$.\Call{push}{$cur$}}
		\EndWhile
		\State
		\While {\textbf{not} $deleteQueue$.\Call{empty}{\null}}
			\State {$cur$ $\gets$ $deleteQueue$.\Call{front}{\null}}
			\State {$deleteQueue$.\Call{pop}{\null}}
		
			\ForEach{$vox$ $\in$ $cur.dll$}
				\State \Call{deleteFromDLL}{$vox.coc$, $vox.pos$}
				\State {$vox.coc$ $\gets$ $\mathcal{IP}$}
				\State {$vox.dis$ $\gets$ $\infty$}
				\ForEach{$nbr \in vox.nbrs$}
					\If {$nbr.coc$ still existing,
					
					$\land$ \Call{dist}{$nbr.coc$, $vox$} $<cur.dis$}
						\State {$vox.dis \gets$ \Call{dist}{$nbr.coc$, $vox$}}
						\State {$vox.coc \gets$ $nbr.coc$}
					\EndIf
				\EndFor
				\If {$vox.coc = $ $\mathcal{IP}$}
					\State {\Call{insertIntoDLL}{$\mathcal{IP}$, $vox.pos$}}
				\Else
					\State {\Call{insertIntoDLL}{$vox.coc$, $vox.pos$}}
					\State {$updateQueue$.\Call{push}{$vox$}}
				\EndIf
			\EndFor
		\EndWhile
	\end{algorithmic}
\end{algorithm}

\subsection{ESDF Updating Algorithm for Limited Observations}
\label{sec:NewlyObserved}
One thing we ignored in the previous two sections is the influence of limited observations. Voxels newly observed may only update its neighbors, without updated by previous existing voxels, which will make the whole system inconsistent. For example, in a 1-D case, only an obstacle at $(0)$ is observed at the beginning. In the next occupancy integration, $(1)-(3)$ are observed and $(3)$ is an obstacle. After executing the Alg.~\ref{alg:Framework}, the ESDF of $(1)$ will be updated to 2, which should be 1 from obstacle $(0)$. This happens because in Alg.~\ref{alg:Framework}, ESDF of the newly observed ones may not be updated from previous existing voxels. So we need to modify the algorithm, to add Alg.~\ref{alg:Patch} between Line 3-4 in Alg.~\ref{alg:Framework}.

\begin{algorithm}[htb]
	\caption{  Patch Code for Limited Observations }
	\label{alg:Patch}
	\begin{algorithmic}[1]
		\Require
		Put these code between Line 3-4 in Alg.~\ref{alg:Framework}
		\Ensure
		The whole map is updated accurately even with limited observations
		\State {$flag \gets \textbf{false}$ } 
		\ForEach{$nbr \in cur.nbrs$}
			\If {\Call{dist}{$nbr.coc, cur.pos$} $< cur.dis$}
				\State {$cur$.$dis$ $\gets$ \Call{dist}{$nbr.coc$, $cur.pos$}}
				\State {\Call{deleteFromDLL}{$cur.coc$, $cur.pos$}}
				\State {$cur.coc$ $\gets$ $nbr.coc$}
				\State {\Call{insertIntoDLL}{$cur.coc$, $cur.pos$}}
				\State {$flag \gets \textbf{true}$ } 
			\EndIf
		\EndFor
		
		\If {$flag$}
			\State {$updateQueue$.\Call{push}{$cur$}}
			\State {\textbf{continue}}
		\EndIf
	\end{algorithmic}
\end{algorithm}

\subsection{Theoretical Analysis}

\label{sec:Optimality}
\subsubsection{Optimality}
Our algorithm is equivalent to update ESDF value of a voxel by the shortest Euclidean distance between this voxel and closest obstacles of its neighbors. The ESDF value is always an "Euclidean" distance. It is obviously accurate than quasi-Euclidean, because the latter uses the length of the broken line as ESDF value.  However, no matter which connectivity we choose, ESDF updating algorithm based on BFS cannot be exactly accurate. We will prove this conclusion below, and gives the relationship between connectivity and optimality.

\begin{figure}[t]
	\centering
	\subfigure[\label{fig:con-4} 2D case with 4-connectivity ]
	{\includegraphics[width=0.43\columnwidth]{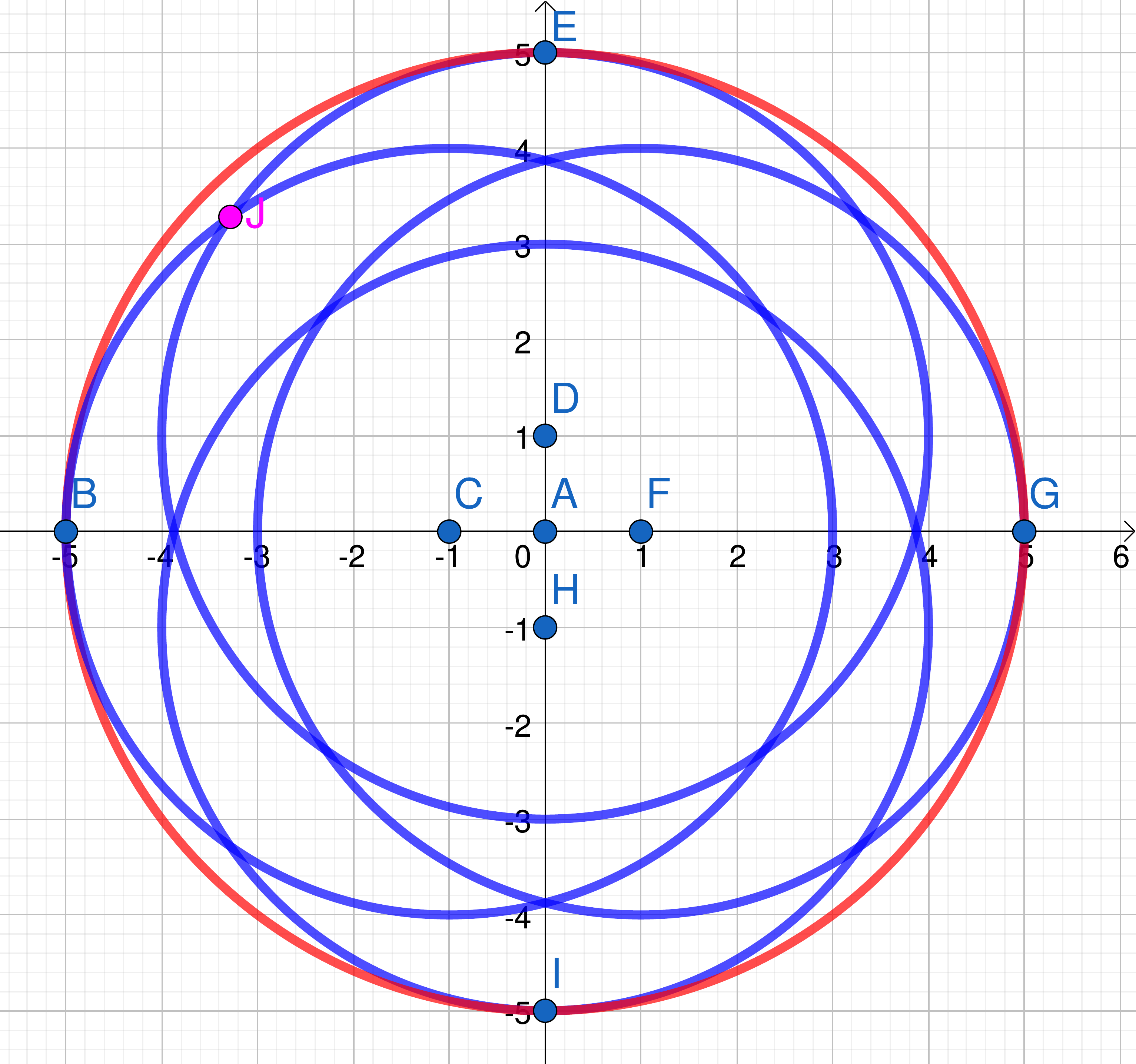}}
	\subfigure[\label{fig:con-8} 2D case with 8-connectivity ]
	{\includegraphics[width=0.43\columnwidth]{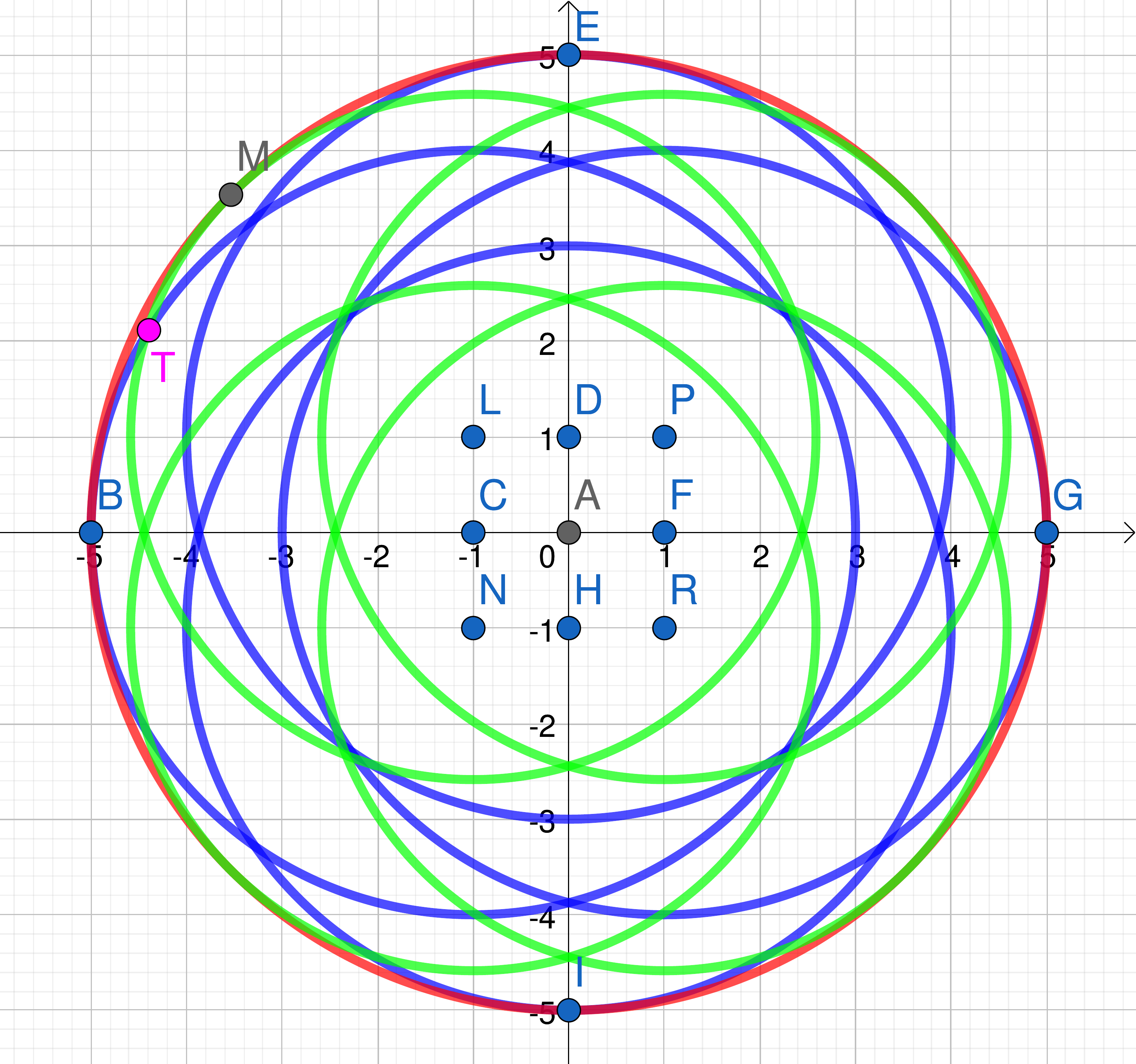}}
	\subfigure[\label{fig:con-4-closer} Making obstacle B (the leftmost point) closer to origin ]
	{\includegraphics[width=0.43\columnwidth]{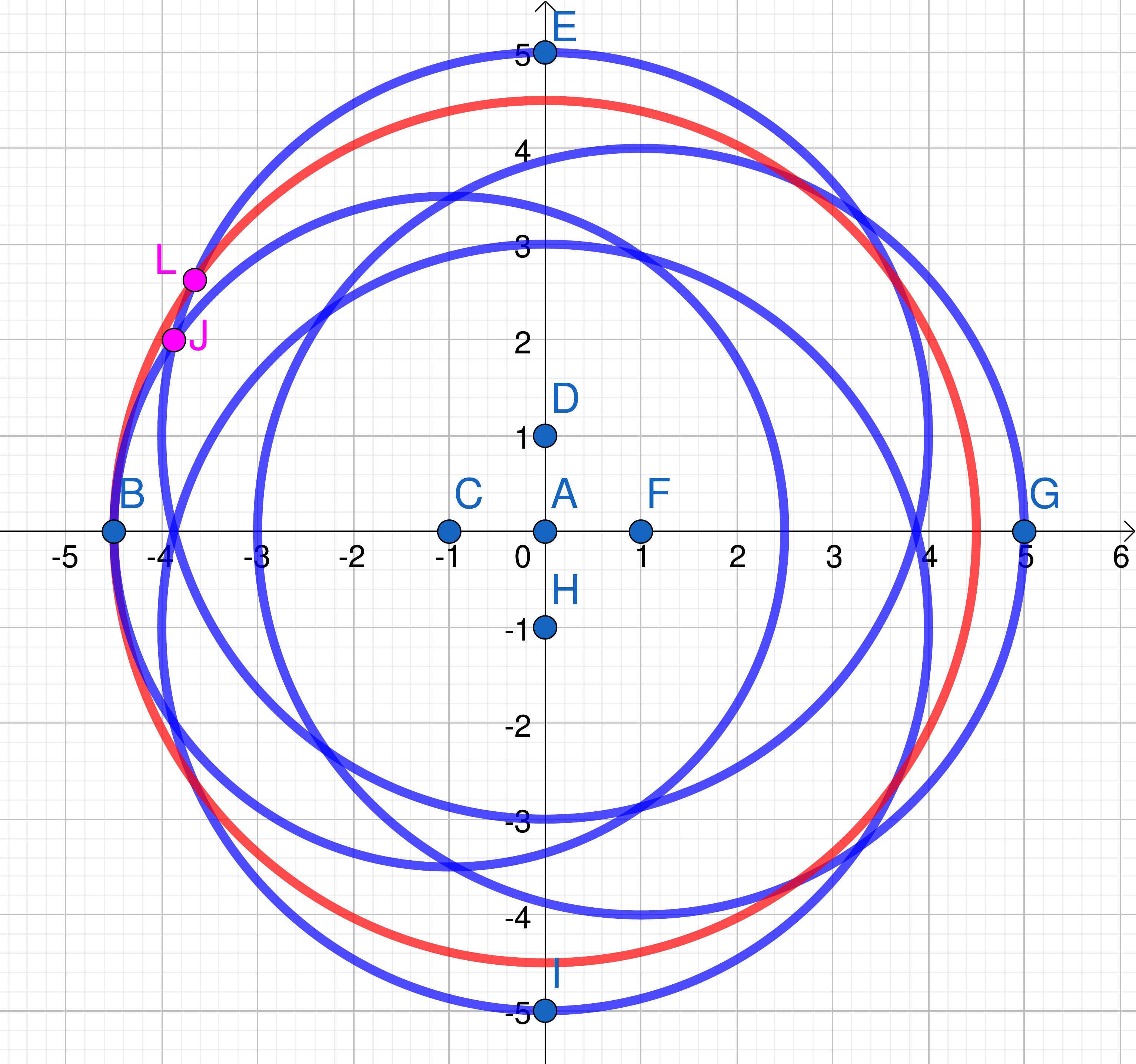}}
	\subfigure[\label{fig:con-4-farther} Making obstacle B (the leftmost point) farther to origin ]
	{\includegraphics[width=0.43\columnwidth]{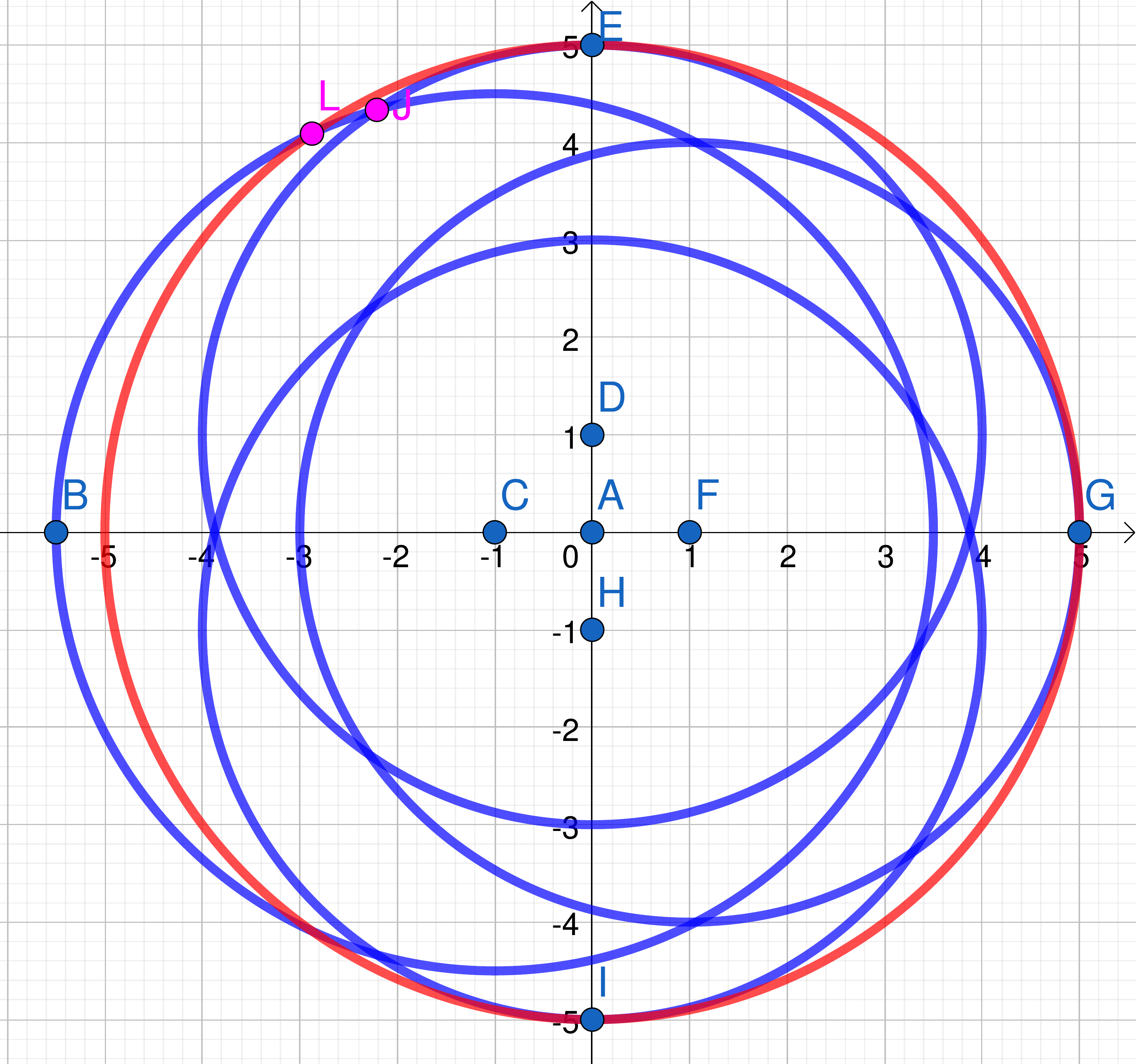}}
	\caption{\label{fig:circles} Circles Illustration for the Optimality Analysis}
\end{figure}

Think about a 2D case whose connectivity is 4. The point need to be updated is the origin, where its neighbors are $nbrs=\{(0, 1), (0, -1), (1, 0), (-1, 0)\}$. Assume the closest obstacle of theirs are $obs_{nbrs}=\{(0, 5), (0, -5), (5, 0), (-5, 0)\}$ correspondingly. According to the ESDF update algorithm, the distance of origin point will be updated to $5$, with the closest obstacle be one of $obs_{nbrs}$. As shown in the Fig.~\ref{fig:con-4}, the center of the red circle is at the origin and its radius is $5$, centers of blue circles are points in $nbrs$ and their radii are $4$. With the assumption that $obs_{nbrs}$ are the closest obstacles of $nbrs$, there are no obstacles inside any blue circle. However, there still remains space (called remaining space) that is outside the blue circles while inside the red circle. The distance between points in that space and $nbrs$ are farther than the distance between $obs_{nbrs}$ and $nbrs$, while the distance between points in that space and origin are closer than the distance between $obs_{nbrs}$ and origin. Fig.~\ref{fig:con-8} also shows a case whose connectivity is 8. In fact, no matter which connectivity we choose, this algorithm cannot be accurate, because using finite small circles inside a big circle to cover the big circle is impossible, and there always remains space to give counter-examples. 

Actually, this is the worst case for ESDF Updating Algorithm using BFS. If one of $obs_{nbrs}$ moves to another point on or inside its original corresponding blue circle, as shown in Fig.~\ref{fig:con-4-closer}, the closest obstacle of origin will be changed and ESDF of origin will be smaller, which in return makes the red circle and remaining space smaller. If it moves outside its original corresponding blue circle, as shown in Fig.~\ref{fig:con-4-farther}, that blue circle becomes bigger and also makes remaining space smaller. The error comes from the difference between the calculated distance from the algorithm and the actual distance. From the illustration, error decreases as radius increases. With the consideration of few occurrences of the worst cases, and the fact that only integer points inside the remaining space will influence our algorithm, the Root Mean Squared (RMS) Error of our algorithm is acceptable in practice. The experimental results will be shown in Section~\ref{sec:SimulationResults}.

\subsubsection{Time Complexity}
For the Alg.~\ref{alg:covert}, every voxel which is either in $insertQueue$ or whose closest obstacle is in $deleteQueue$ is handled only once. If we use an FIFO queue, the time complexity is $\Theta(k)$, where $k$ is the number of all necessary voxels needed to be handled.

And for the Alg.~\ref{alg:Framework}, if we use a priority queue to do the BFS procedure, we are sure that all voxels are popped from $updatequeue$ and handled only once to update its neighbors. Then the time complexity is $\Theta(n \log n)$, where $n$ is the number of all voxels need to be updated, and the $\log n$ factor comes from the nature of priority queue.

\subsubsection{Space Complexity}
As we described in Section~\ref{sec:indexing}, comparing to the pure occupancy grid map based on a hash table, only VIS is modified for our system. so it is only constant-level higher than pure occupancy grid map. From the data structures we design, we can even make it equal to $\Theta(m)$ if we choose to use $block\_size = 1$, where $m$ is the number of all ever observed voxels, though it is not high-efficient. Thus it is significant to have highly customizable Indexing Data Structures with consideration of requirements of the real system.

\section{Results}
\label{sec:Results}
\subsection{Experiments on Real-world Datasets}
\label{sec:DatasetResults}
\begin{figure}[t]
	\centering
	\includegraphics[width=0.8\columnwidth]{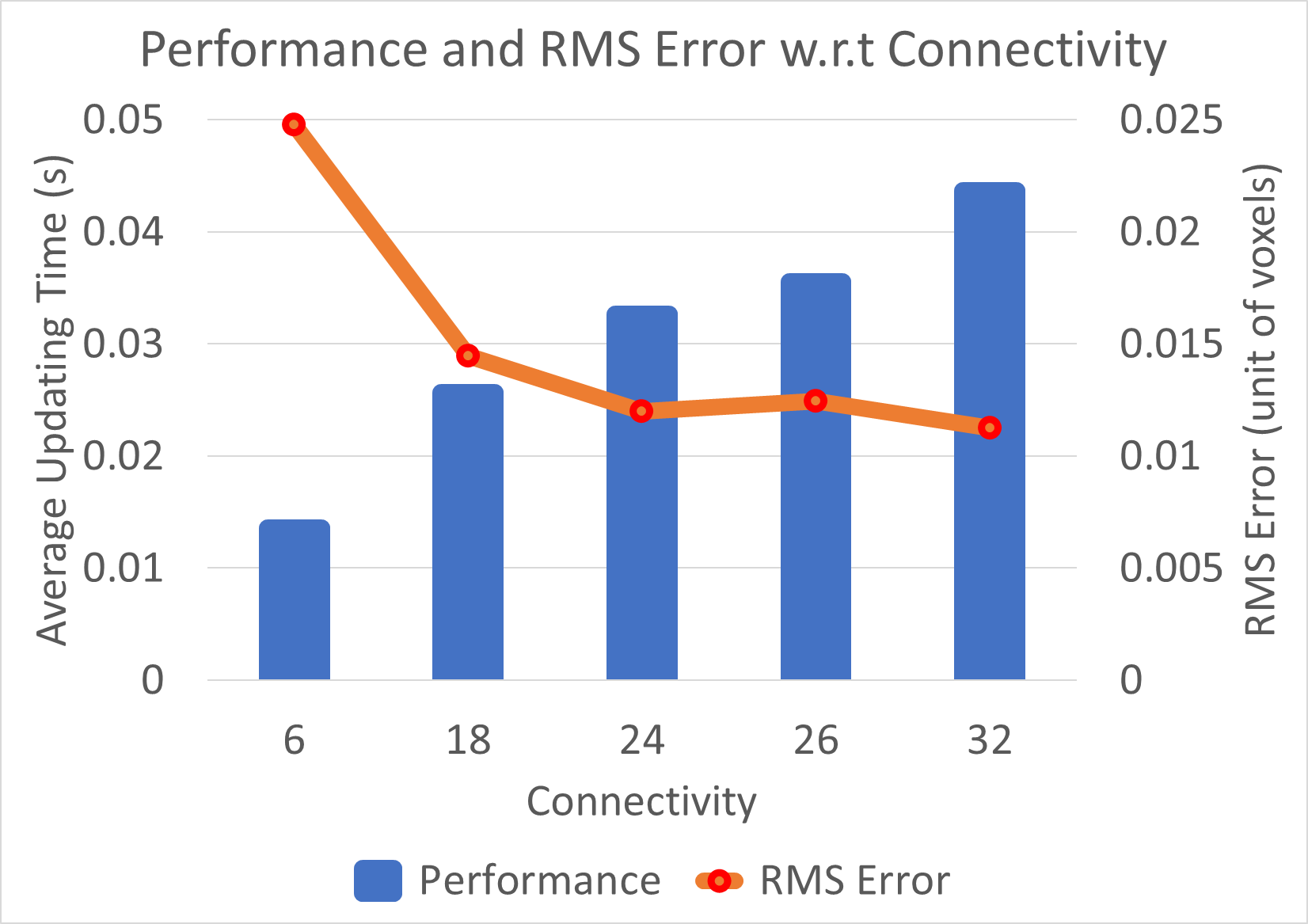}
	\caption{ Time and RMS Error with respect to different Connectivities. The x-axis is the type of connectivity. The left y-axis is the performance measured by average updating time, the right y-axis is the RMS Error in the unit of voxels.
		\label{fig:connectivityresults}}
\end{figure}

\begin{figure}[t]
	\centering
	\includegraphics[width=0.8\columnwidth]{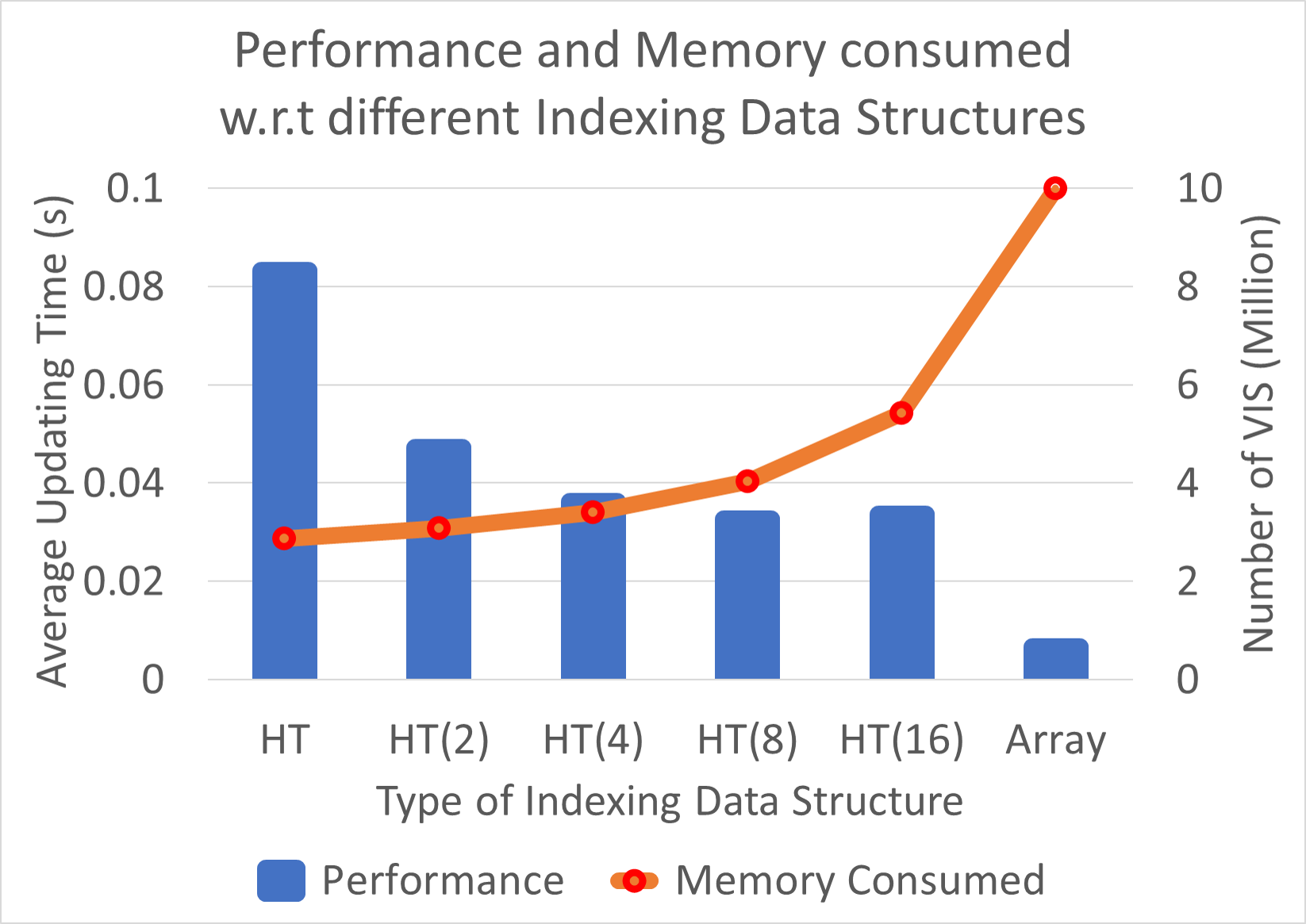}
	\caption{ Time and Space with respect to different Indexing Data Structures. The x-axis is the type of Indexing Data Structure, where HT(n) means Hash Table with $block\_size=n$. The left y-axis is the performance measured by average updating time, the right y-axis is the memory consumption measured by the number of VIS.
		\label{fig:indexingresults}}
	\vspace{-1.0cm}
\end{figure}
In this section, we test different parameters of our system FIESTA, and then we compare it with Voxblox in different level of voxel size. All the experiments in this part are using i7-8700k at 3.7GHz and only one thread is used.

\subsubsection{Parameters Tuning} There are a lot of parameters inside our system, such as Connectivity and which Indexing Data Structure is used. A set of experiments will be provided to test the optimal parameters for our system with respect to the accuracy, performance and memory consumption. Our system asks for ESDF updates at a frequency of every 0.5s, and the performance is measured by the average updating time. The largest memory consumption is VIS, so we use a number of VIS as a measurement of memory consumption. After running the whole dataset, we establish a K-D Tree using the occupancy information obtained from an occupancy grid map and then query for the distance for each voxel inside map as ground truth. After that Root Mean Squared (RMS) Error is used in the unit of voxels to demonstrate the scale of error. In this section, Cow\_and\_Lady dataset\footnote[3]{\url{https://projects.asl.ethz.ch/datasets/doku.php?id=iros2017}}\cite{oleynikova2017voxblox} with an RGB-D camera, which contains point cloud data and pose information, is used as the dataset.

First, we fix the Indexing Data Structure to be a hash table with $block\_size = 8$ and test accuracy with respect to connectivity. We try 6-(faces), 18-(faces and edges), 26-(faces, edges, and corners), 24-(faces, edges and 2-step faces, i.e. Manhattan distance less than or equal to 2) and 32-(combine of previous two) connectivities. More connectivity results in more neighbors to be updated therefore gives a higher accuracy with worse computational performance. Thus we need to choose one trade-off in both performance and accuracy. Fig.~\ref{fig:connectivityresults} shows the result. From the illustration, 24-connectivity is one of the best choices for both RMS Error and performance. Then, different Indexing Data Structures are compared. As shown in the Section~\ref{sec:indexing}, we give a lot of choices to provide a trade-off between performance and memory consumption. We fix the connectivity to 24, and then test array-implemented and hash-table-implemented with $block\_size = 1, 2, 4, 8, 16$. Fig.~\ref{fig:indexingresults} shows the result. According to the result, if the map boundary is unknown, a hash table with $block\_size = 8$ is one of the best choices for both time and space complexity. Otherwise, the array has the best performance.

\begin{figure}[t]
	\centering
	\subfigure[\label{fig:grid_map} Cow\_and\_Lady Dataset ]
	{\includegraphics[width=0.49\columnwidth]{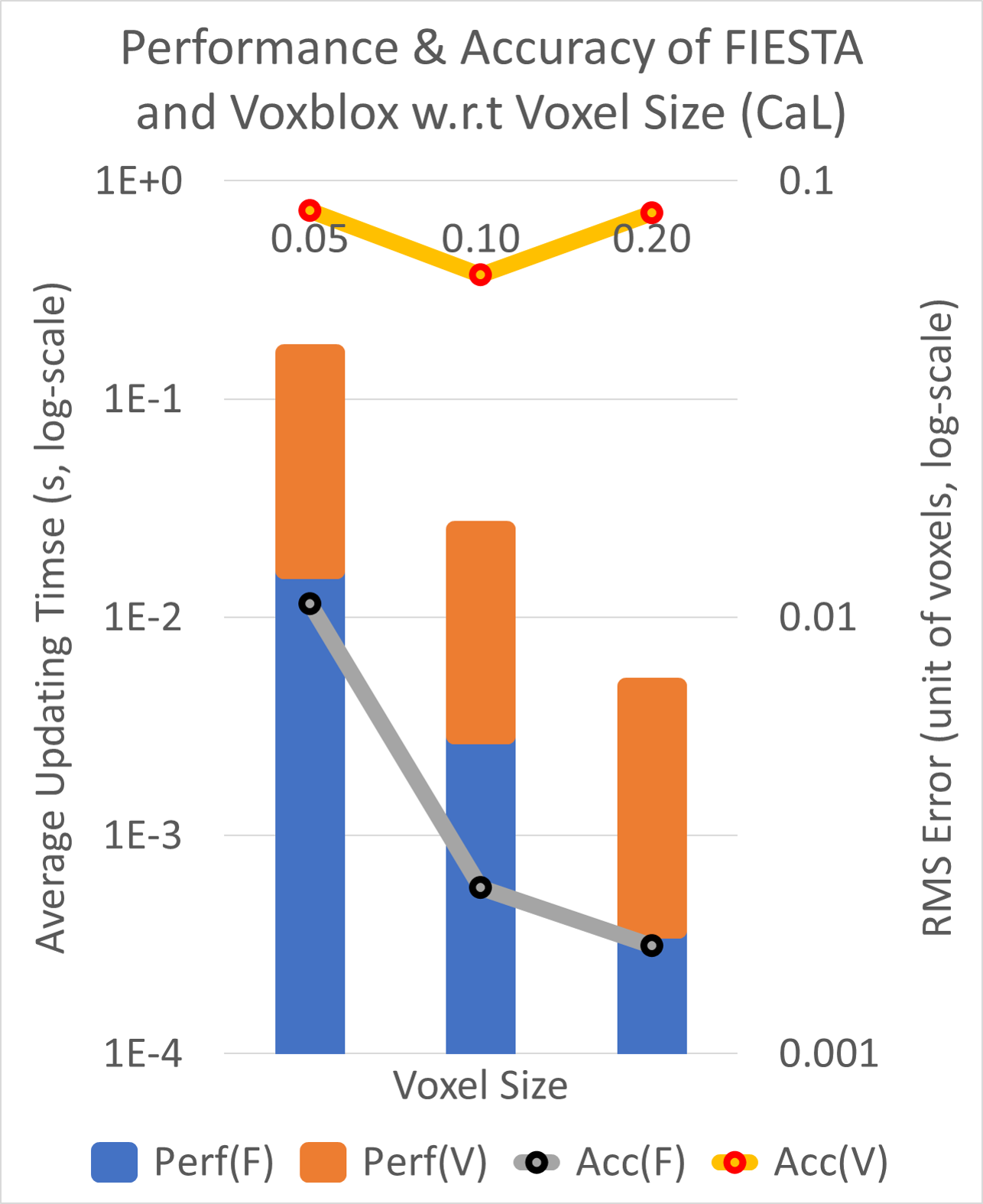}}
	\subfigure[\label{fig:velocity_f} EuRoC Dataset ]
	{\includegraphics[width=0.49\columnwidth]{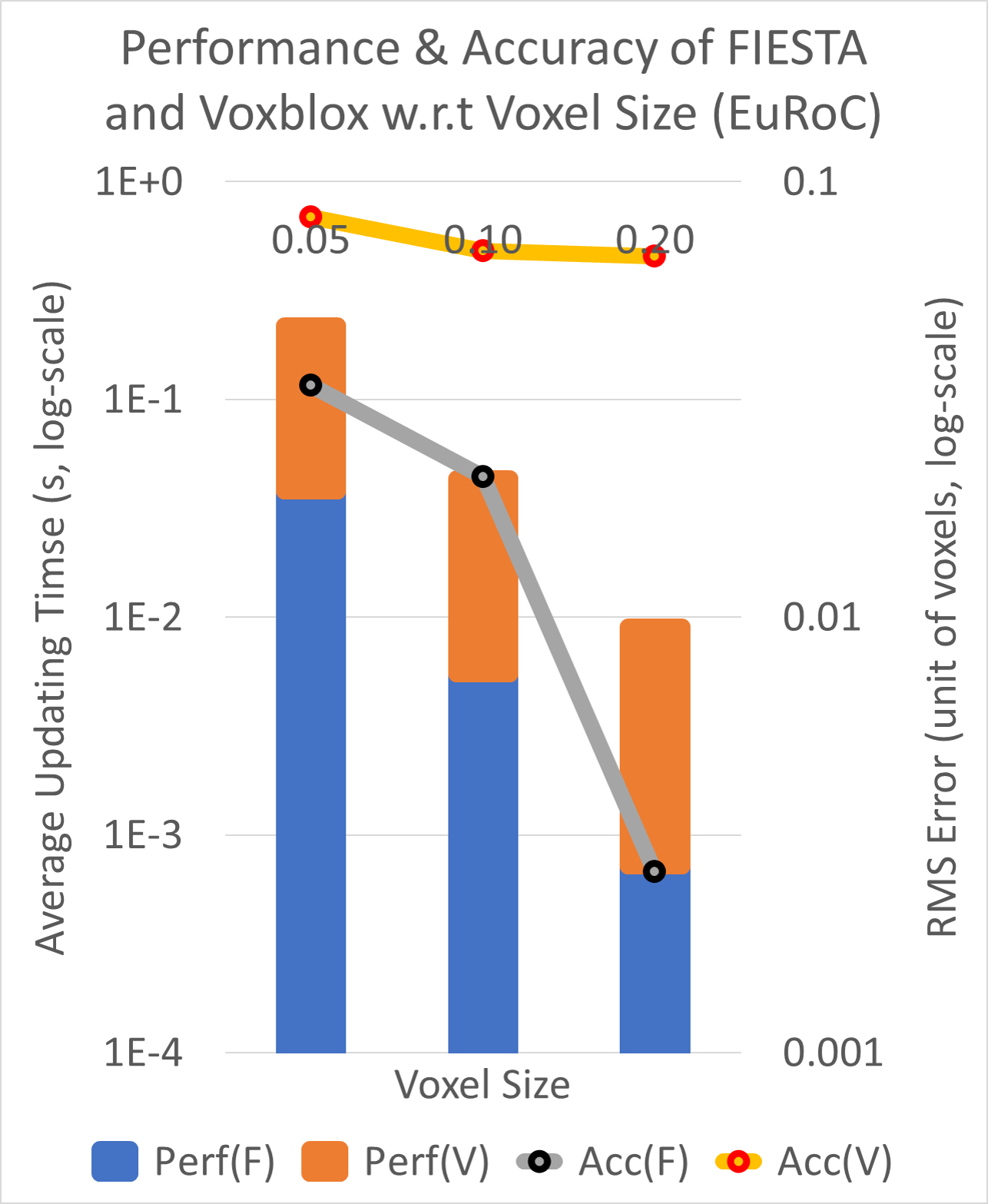}}
	\caption{\label{fig:cal} Performance(Perf) and Accuracy(Acc) between FIESTA(F) and Voxblox(V) with respect to Voxel Size compared using Cow\_and\_Lady and EuRoC Datasets. The x-axis is voxel size. The left y-axis is the performance measured by average updating time, the right y-axis is the RMS Error in the unit of voxels.}
	\vspace{-0.9cm}
\end{figure} 

\subsubsection{Comparison with Voxblox} 
\begin{figure}[t]
	\centering
	\subfigure[\label{fig:grid_map} Occupancy Grid Map ]
	{\includegraphics[width=0.49\columnwidth]{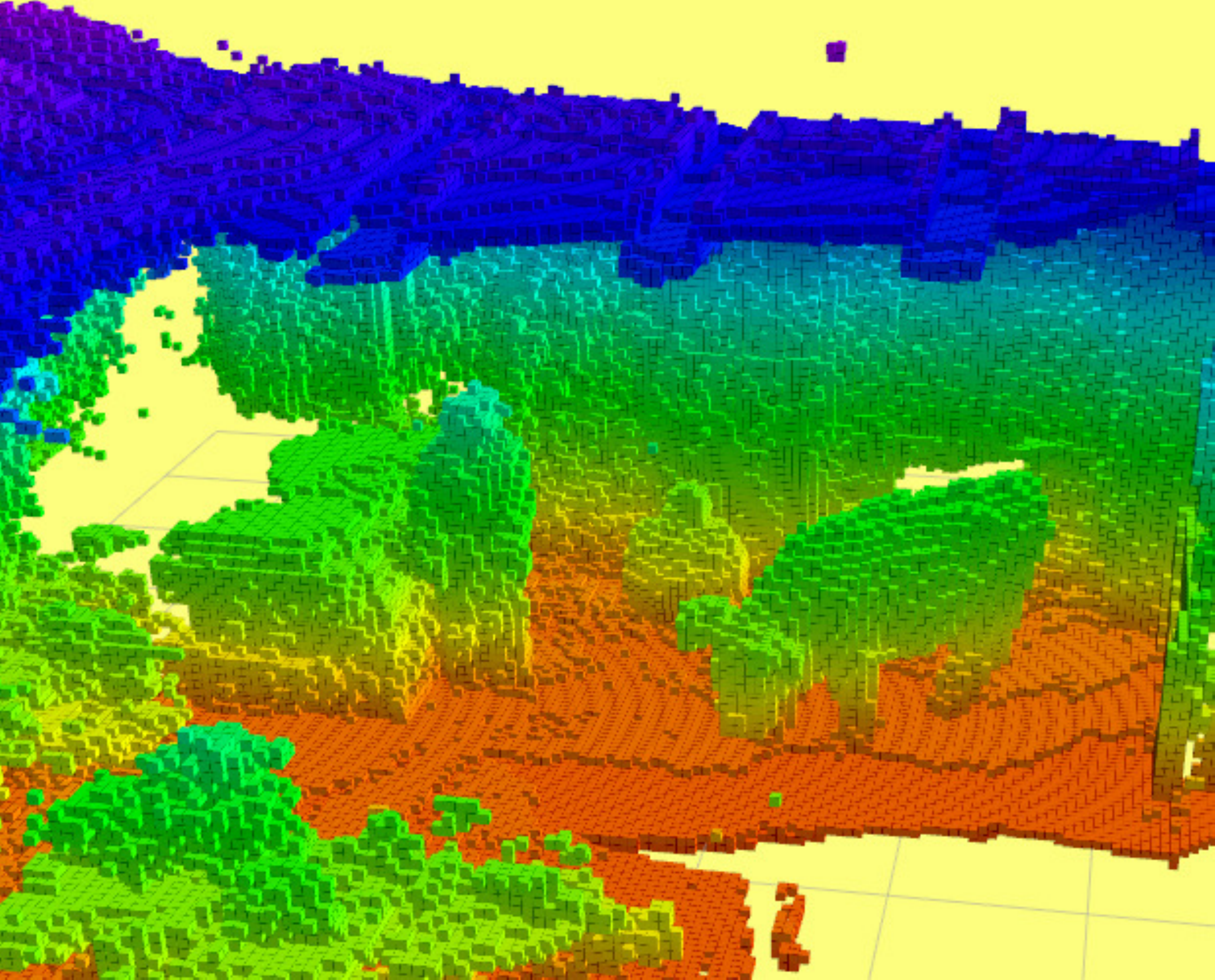}}
	\subfigure[\label{fig:velocity_f} Slice of ESDF Map ]
	{\includegraphics[width=0.49\columnwidth]{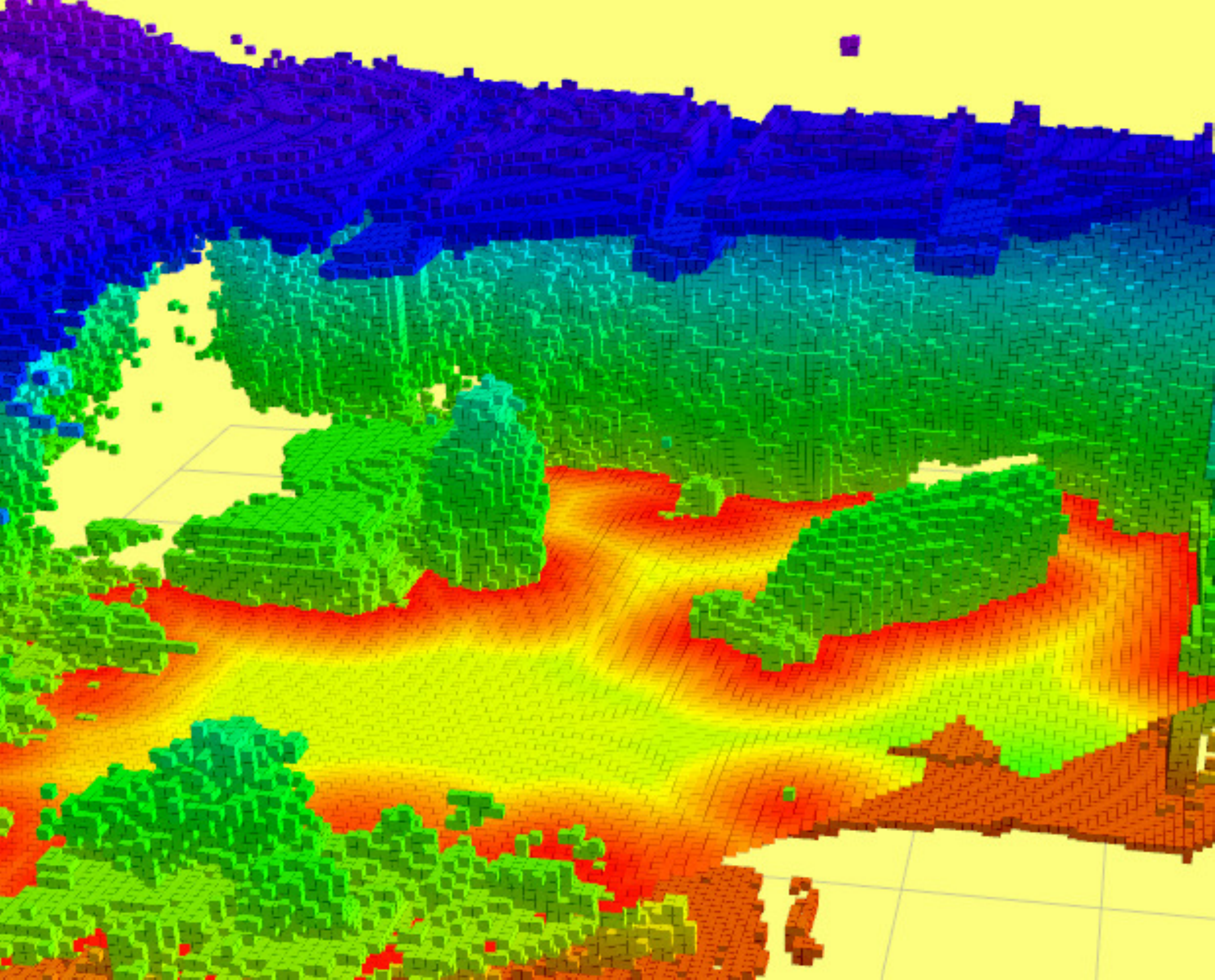}}
	\caption{\label{fig:comparison} Visulization of Occupancy Grid Map and a slice of ESDF Map using our system with $voxel\_size = 0.05$, running Cow\_and\_Lady Dataset }
	\vspace{-1.2cm}
\end{figure}

A comparison between FIESTA and Voxblox in term of performance and accuracy in different voxel size is provided. Cow\_and\_Lady dataset with an RGB-D camera and EuRoC dataset with a stereo camera are used as datasets for this comparison. We choose to use a hash table with $block\_size = 8$ as the Indexing Data Structure and 24-Connectivity for this comparison. Fig.~\ref{fig:comparison} shows the results, it is clear that our system outperforms Voxblox in an order of magnitude in both performance and accuracy. Visualization of Occupancy Grid Map and a slice of ESDF Map using our system with $voxel\_size = 0.05$, running Cow\_and\_Lady dataset is shown in Fig.~\ref{fig:cal}.

\subsection{Tests of Quadrotor Motion Planning in Simulation}
\label{sec:SimulationResults}
\begin{figure*}[!t]
	\centering
	\subfigure[] {\includegraphics[width=0.47\columnwidth]{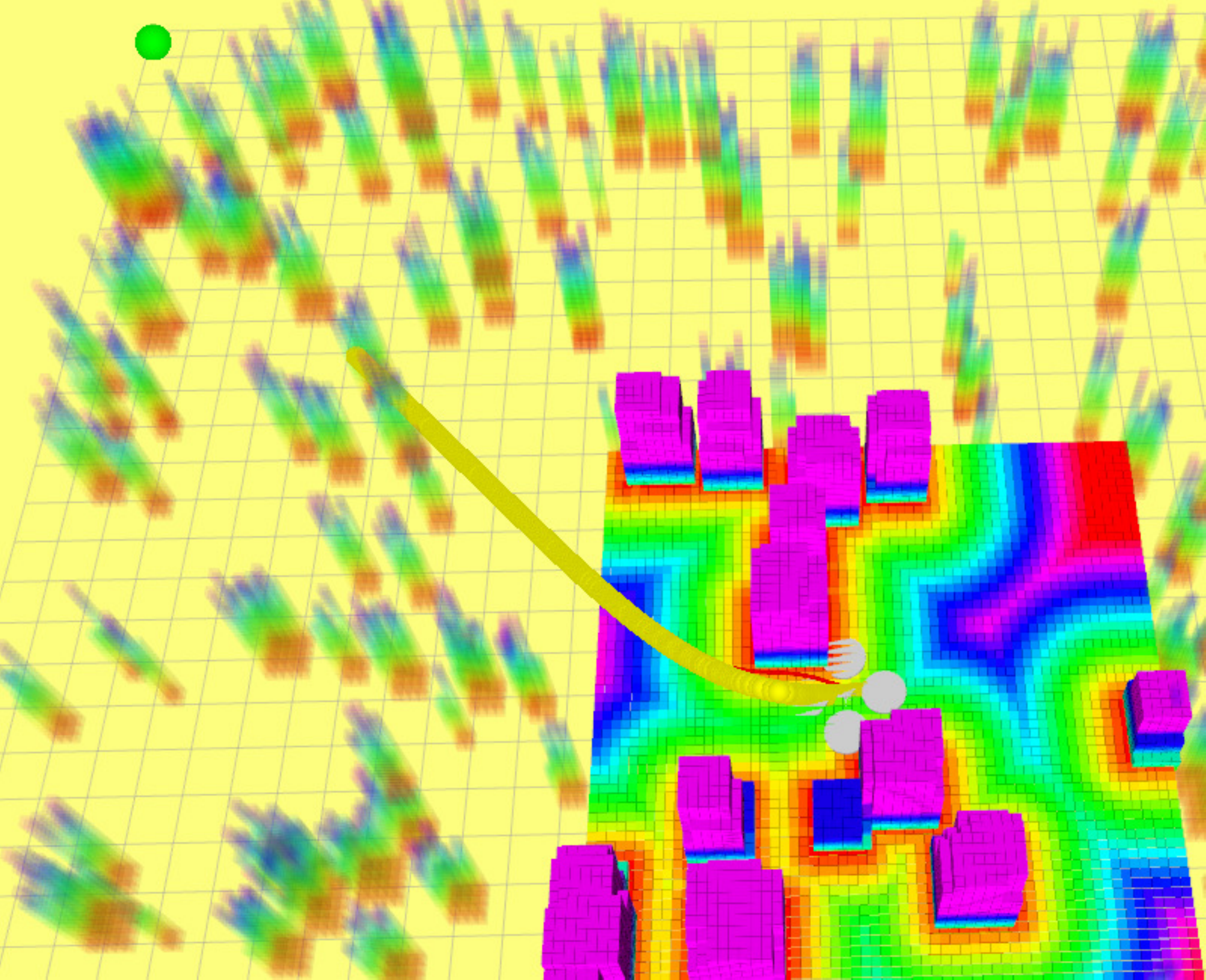}}
	\subfigure[] {\includegraphics[width=0.47\columnwidth]{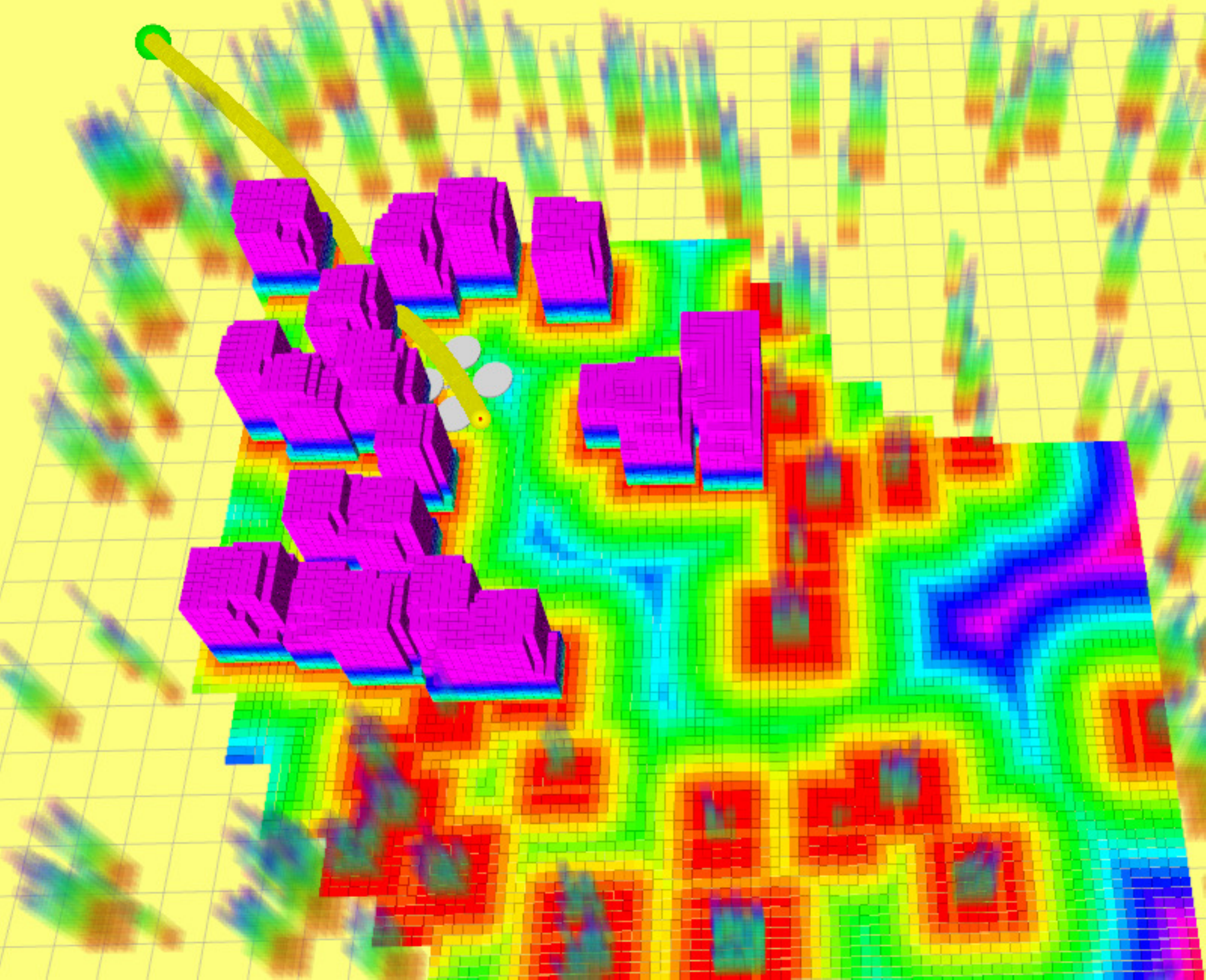}}
	\subfigure[] {\includegraphics[width=0.47\columnwidth]{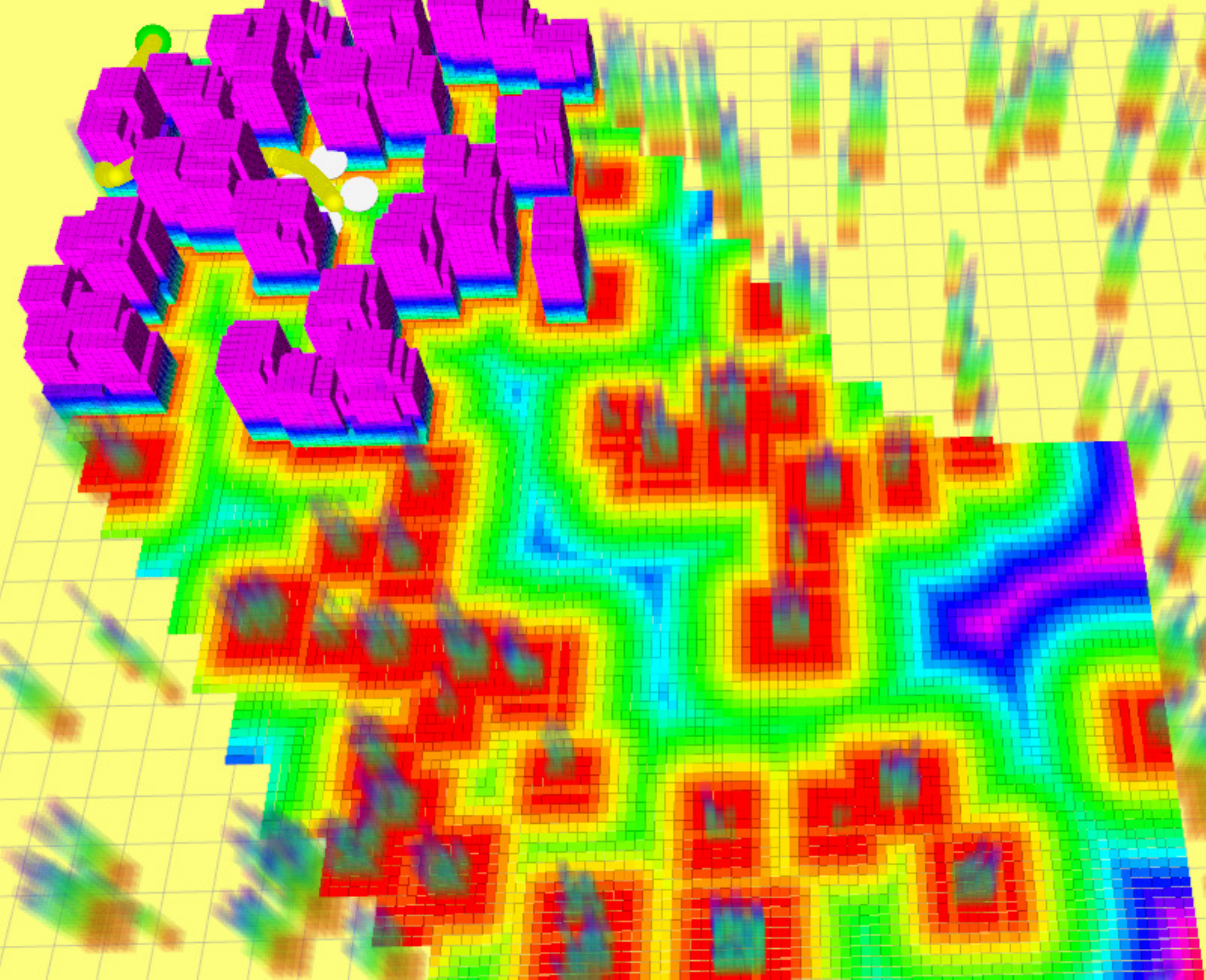}}
	\subfigure[] {\includegraphics[width=0.47\columnwidth]{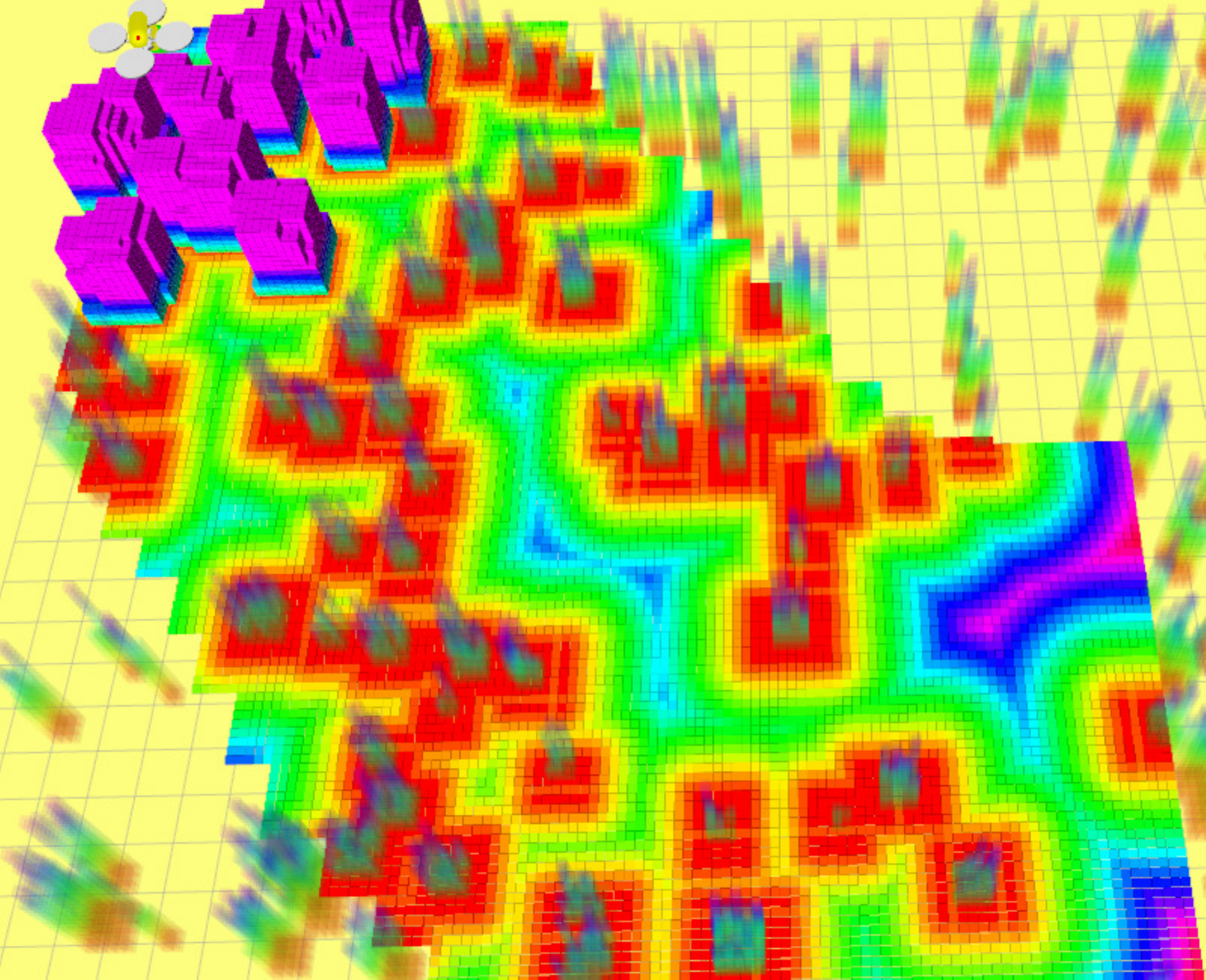}}
	\caption{ Simulation Experiments Results }
	\label{fig:Simulation}
		\vspace{-0.5cm}
\end{figure*}

To prove that our proposed map can be used for motion planning, we present simulated quadrotor flight, where the motion planning method is adopted from~\cite{boyu2019ral}. In simulation, a map as well as the starting and destination points of the MAV are generated randomly. Only obstacles within a radius around the current position of MAV can be sensed. Our system FIESTA builds global ESDF Map incrementally, to help to motion planning algorithm running efficiently. Our sample results are shown in Fig.~\ref{fig:Simulation}.

\subsection{Autonomous Quadrotor Onboard Experiments}
\label{sec:OnboardResults}
\begin{figure*}[!t]
	\centering
	\subfigure[] {\includegraphics[width=0.47\columnwidth,height=2.3in]{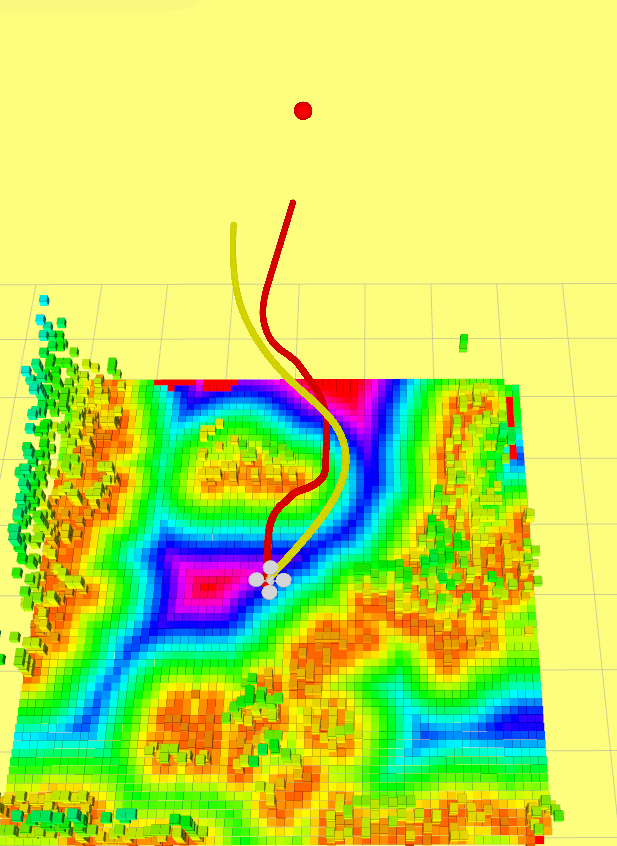}}
	\subfigure[] {\includegraphics[width=0.47\columnwidth,height=2.3in]{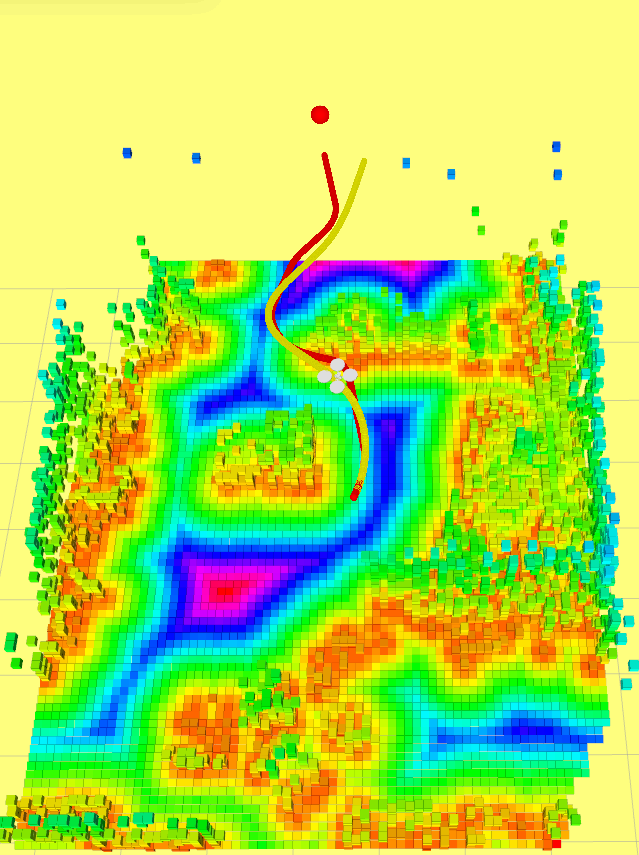}}
	\subfigure[] {\includegraphics[width=0.47\columnwidth,height=2.3in]{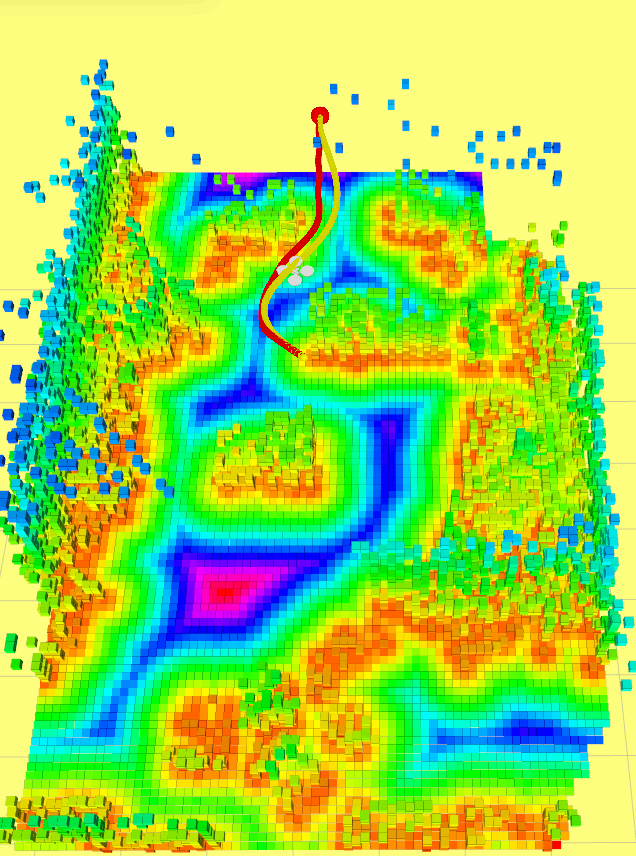}}
	\subfigure[] {\includegraphics[width=0.47\columnwidth,height=2.3in]{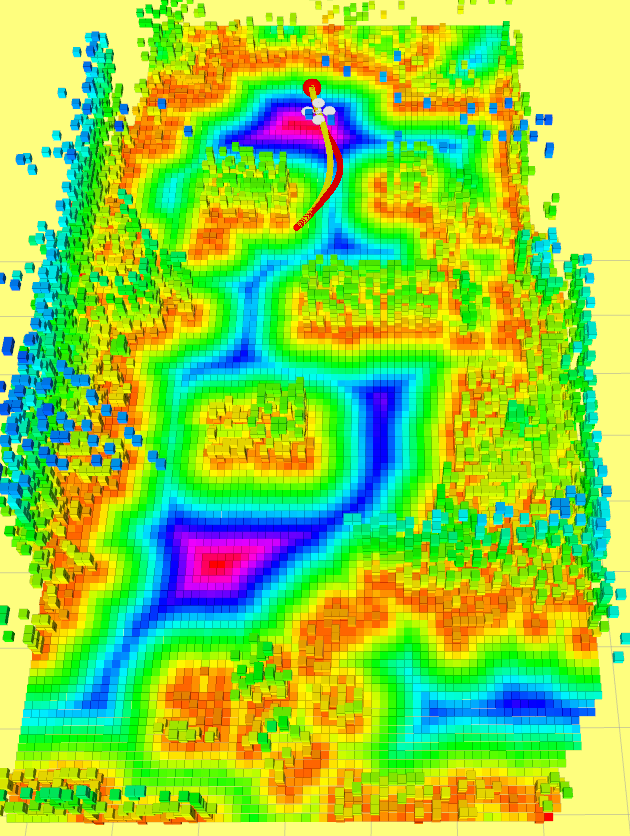}}
	\caption{ Onboard Experiments Results }
	\label{fig:Onboard}
		\vspace{-0.5cm}
\end{figure*}

Onboard experiments are conducted in unknown cluttered environments using the same motion planning approach as in Section~\ref{sec:SimulationResults}. The platform we use is a quadrotor equipped with a Velodyne VLP-16 3D Lidar, which is used for both pose estimation and depth measurement. All state estimation, motion planning, control, and our proposed mapping system are running onboard on a dual-core 3.00GHz Intel i7-5500U processor. Although our ESDF map can finish an update within 20ms,  we update ESDF at 20Hz, due to the low requirement of the flight speed in our experiments. Also, updating ESDF map at a relatively low frequency saves a lot of onboard recourses for other modules such as the planning and localization. Snapshots of the experiment and the incremental mapping and planning results are given in Figs.~\ref{fig:realworld}~and~\ref{fig:Onboard}.

\section{Conclusion}
\label{sec:Conclusion}
In this paper, a lightweight and flexible mapping framework, FIESTA, was proposed. FIESTA builds a global ESDF map incrementally, which is important to motion planning. By introducing two independent updating queues for inserting and deleting obstacles separately, and using Indexing Data Structures and Doubly Linked Lists to maintain voxels, our algorithm updates as few as possible nodes using a BFS framework. Both theoretical and practical analysis in term of time and space complexity, accuracy and optimality are given in this paper. We proved that the updating rule of our method outperforms quasi-Euclidean one for Voxblox. In practical, using a hash table with $block\_size = 8$ and 24-Connectivity, our system beats Voxblox in an order of magnitude in term of accuracy and time complexity.
The proposed ESDF map is also integrated into a completed quadrotor system. With the simulation and onboard experiments, we validate our system by building global ESDF map incrementally high-efficiently.
\bibliography{iros2019luxin} 
\end{document}